\pdfoutput=1
\documentclass[11pt]{article}

\usepackage[final]{acl}
\usepackage{times}
\usepackage{latexsym}
\usepackage{lipsum}  

\usepackage[T1]{fontenc}
\usepackage[utf8]{inputenc}

\usepackage{microtype}

\usepackage{inconsolata}

\usepackage{url}            % simple URL typesetting
\usepackage{hyperref}       % hyperlinks
\usepackage{booktabs}       % professional-quality tables
\usepackage{amsfonts}       % blackboard math symbols
\usepackage{nicefrac}       % compact symbols for 1/2, etc.
\usepackage{microtype}      % microtypography
\usepackage{xcolor}         % colors
\usepackage{enumitem}
\usepackage{graphicx}
\usepackage{float}
\usepackage{pifont}
\usepackage{amssymb}
\usepackage{longtable}
\usepackage{algorithm}
\usepackage{algpseudocode}
\usepackage{amsmath}
\usepackage{geometry}
\usepackage{subcaption}
\usepackage[normalem]{ulem}
\usepackage{titletoc}
\usepackage{tabularx} % Add this to your preamble
\usepackage{adjustbox}
\usepackage{makecell}
\usepackage[table]{xcolor}
\usepackage{arydshln}
\usepackage{booktabs}
\usepackage{multirow}
\usepackage{hyperref}
\usepackage{enumitem}
\usepackage{float}

\newcommand\benchmarkName{\texttt{\textbf{TechING}}}
\newcommand\modelName{\texttt{\textbf{LLama-VL-TUG}}}
\newcommand\imageCountNumber{115,014}
\newcommand\imageCount{$\imageCountNumber$}
\newcommand\imageCountReal{$545$}

\title{\benchmarkName: Towards Real World Technical Image Understanding via VLMs}

\author{
  \textbf{Tafazzul Nadeem\thanks{Equal contribution}}$^{\dagger}$ \quad
  \textbf{Bhavik Shangari\footnotemark[1]}$^{\ddagger}$ \quad
  \textbf{Manish Rai}$^{\ddagger}$ \\
  \textbf{Gagan Raj Gupta}$^{\ddagger}$ \quad
  \textbf{Ashutosh Modi}$^{\dagger}$ \\
  $^{\dagger}$IIT Kanpur \quad $^{\ddagger}$IIT Bhilai \\
  \texttt{\{tafazzul23, ashutoshm\}@cse.iitk.ac.in}, \\
  \texttt{\{bhaviks, manishr, gagan\}@iitbhilai.ac.in}
}

\begin{document}
\maketitle

% \begin{abstract}
% This document is a supplement to the general instructions for *ACL authors. It contains instructions for using the \LaTeX{} style files for ACL conferences. 
% The document itself conforms to its own specifications, and is therefore an example of what your manuscript should look like.
% These instructions should be used both for papers submitted for review and for final versions of accepted papers.
% \end{abstract}

\begin{abstract}
Professionals working in technical domain typically hand-draw (on whiteboard, paper, etc.) technical diagrams (e.g., flowcharts, block diagrams, etc.) during discussions; however, if they want to edit these later, it needs to be drawn from scratch. Modern day VLMs have made tremendous progress in image understanding but they struggle when it comes to understanding technical diagrams. One way to overcome this problem is to fine-tune on real world hand-drawn images, but it is not practically possible to generate large number of such images. In this paper, we introduce a large synthetically generated corpus (reflective of real world images) for training VLMs and subsequently evaluate VLMs on a smaller corpus of hand-drawn images (with the help of humans). We introduce several new self-supervision tasks for training and perform extensive experiments with various baseline models and fine-tune Llama 3.2 11B-instruct model on synthetic images on these tasks to obtain \modelName, which significantly improves the \textbf{ROUGE-L} performance of Llama 3.2 11B-instruct by \textbf{2.14x} and achieves the best all-round performance across all baseline models. On real-world images, human evaluation reveals that we \textbf{achieve minimum compilation errors} across all baselines in 7 out of 8 diagram types and improve the average \textbf{F1 score} of Llama 3.2 11B-instruct by \textbf{6.97x}.
\end{abstract}

%\vspace{-3.5mm}
\section{Introduction} \label{sec:intro}
%\vspace{-2mm}

\begin{figure*}[t]
            \centering
            \includegraphics[width=\linewidth]{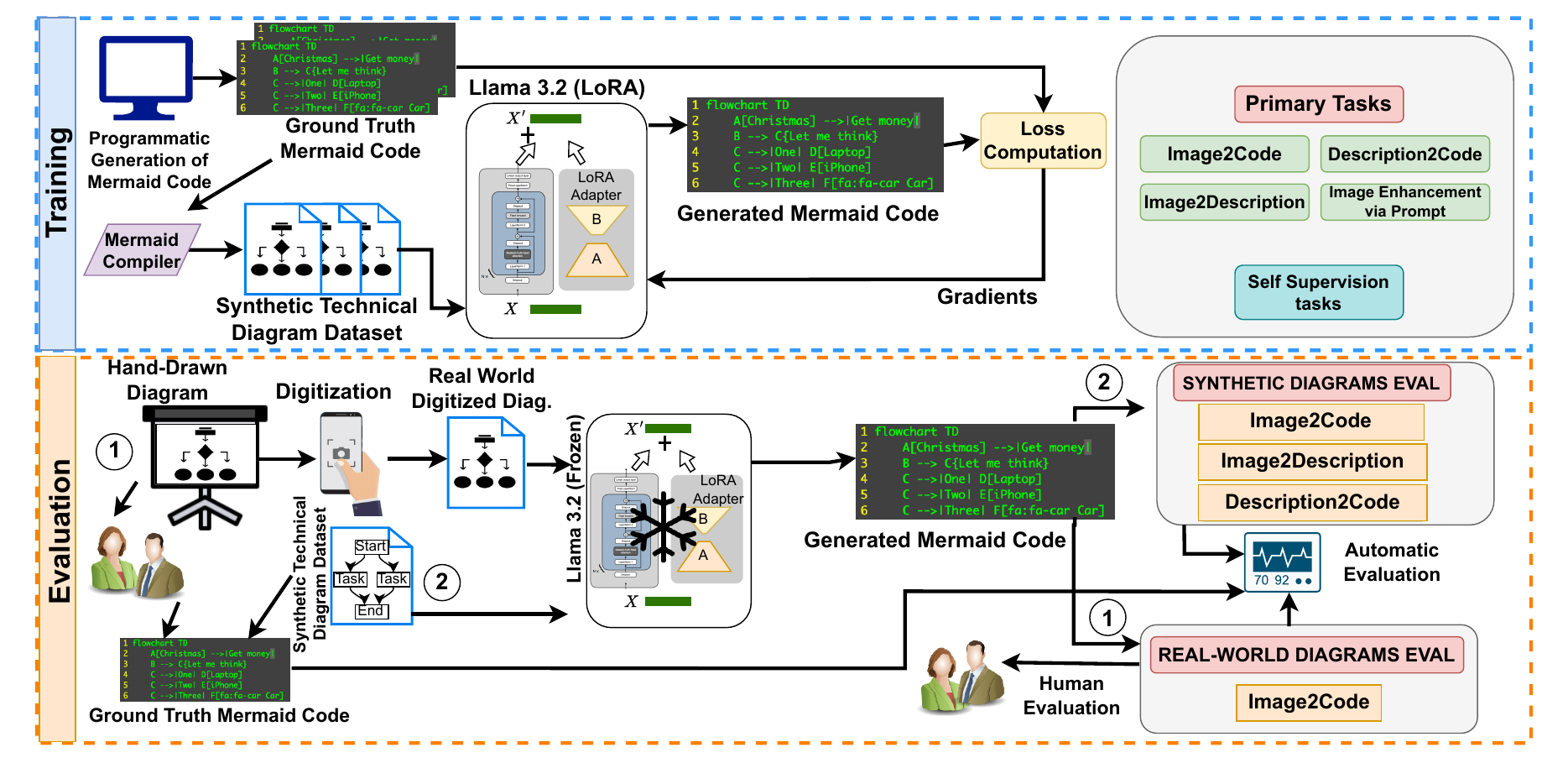}
            \vspace{-7mm}
            \caption{Figure illustrating the pipeline for technical diagram understanding}
            \label{fig:entire-pipline}
            %\vspace{-5mm}
\end{figure*}

Technical discussions typically involve ideation and translating those ideas into hand-drawn diagrams (e.g., flowcharts, class, packet diagrams) on a whiteboard or paper. Such diagrams are often digitized in the form of images in jpg/png formats for documentation purposes, with contents of the image not readily machine-understandable, making it challenging to edit these later via natural language queries. Consequently, we need techniques that enable computers to understand digitized images of technical diagrams, making it feasible to edit these via natural language queries.  

\noindent Recently, Vision Language Models (VLMs) have shown impressive performance on various benchmarks \citep{yue2024mmmu,liu2024mmbench}, however, one area that remains under-explored is an assessment of the abilities of VLMs with regard to technical diagram understanding and generation. Our initial experiments (\S\ref{sec:results}) reveal that many of the pre-trained VLMs have limited understanding of technical diagrams. 

Many commercial tools are being developed (e.g. \url{www.eraser.io}, \url{www.whimsical.com}, \url{www.notegpt.io}, \url{www.diagrammingai.com}, \url{www.visily.ai}) with the aim of helping users generate technical diagrams via prompts. These tools have limited performance (based on our initial study) and focus mainly on system diagrams or workflow-related images. There is a need for models that can understand technical images, as these can be beneficial for various applications such as diagram editing, accelerating design iterations, enhancing collaboration and versioning, automation of document workflows, requirements-driven modeling, and diagram creation. 

\noindent One possible way to address this gap is to develop domain-specific VLMs that can understand multiple types of technical diagrams. However, this comes with several challenges: \textbf{1)} a lacuna of corpus of technical diagrams that are reflective of real-world settings, particularly hand-drawn technical diagrams. \textbf{2)} Technical diagrams are domain-specific and image types vary across domains by a large margin; therefore, the model should be able to adapt to and generalize the domain. For example, \textit{class diagrams}  are used in software documentation to represent the system design, and \textit{packet diagrams} are used for standardizing the packet structures of various protocols in network specifications such as TCP/IP; both types of diagram are very different (see App. \ref{app:sec:image-examples}). %Fig. \ref{fig:image_examples1} and Fig. \ref{fig:image_examples2}).
\textbf{3)} Technical diagrams pose representational challenges due to the tight coupling between geometric shapes (e.g., boxes, circles, arrows) and the text within or on these geometrical shapes (e.g., text on an arrow, text inside a box, etc.), requiring models to process text and geometrical shapes jointly. \textbf{4)} Hand-drawn technical diagrams can have large variations, same diagram can be made by different people in various ways. \textbf{5)} An effective way to build understanding in a VLM is by introducing an intermediate representation (in the form of computer code) that encodes the structure and semantics of drawings in a simplified, symbolic form. However, manually constructing a large corpus of hand-drawn images paired with their corresponding code representations is highly resource-intensive, \textit{leading to a low-resource setting}. To address these challenges, we propose an alternate approach: %, where we may not have enough data to fine-tune VLM. 

\begin{table*}[t]
    \centering
    \small
    %\begin{adjustbox}{width=\textwidth}
    \resizebox{1.0\linewidth}{!}{
    \begin{tabular}{lcccccc}
        \toprule
        \textbf{Corpus} & \textbf{Diagram types} & \makecell{\textbf{Image Dataset} \\ \textbf{Size (English)}} & \makecell{\textbf{Real World} \\ \textbf{Images} \\ \textbf{(Hand Drawn)}}  & \makecell{\textbf{Real World} \\ \textbf{Images} \\ \textbf{(Scraped)}} & \makecell{\textbf{Structural}\\ \textbf{Summary}} & \makecell{\textbf{Diagram} \\ \textbf{Code}} \\
        \midrule
        \textbf{FC Database \citep{inproceedings}}     &     Flowchart      &     78        & 78 &    0     &      \texttimes      &       \texttimes      \\ \midrule
        \textbf{CBD \citep{bhushan-lee-2022-block}}     &     Block      &     502        & 0 &    502     &      \texttimes      &       \texttimes      \\ \midrule %\hdashline
        \textbf{FlowchartQA \citep{tannert-etal-2023-flowchartqa}}          &      Flowchart       &       \textasciitilde1,000,000      &  0 &     166      &     \texttimes      &      \texttimes     \\ \midrule %\hdashline
        \textbf{BD-EnKo \citep{bhushan-etal-2024-unveiling}}      &      \makecell{Flowchart, Graph, \\ Journey, Sequence, \\ State, C4}       &    46,599         &   0 &    91      &       \texttimes      &       \texttimes       \\ \midrule %\hdashline
        \textbf{FlowVQA \citep{singh-etal-2024-flowvqa}}          &       Flowchart      &       2,272    & 0 & 0 &    \texttimes       &    \texttimes         \\ \midrule %\hdashline
        \textbf{\citet{ai-etal-2024-advancement}}      &      \makecell{Knowledge Graph,\\ Route Map \\Flowchart,\\ Mind Map,\\Gantt Chart}       &      1,314       &  0 &  1,314   &          \texttimes    &       \texttimes      \\ \midrule %\hdashline

        \textbf{FlowLearn \cite{pan2024flowlearnevaluatinglargevisionlanguage}}          &      Flowchart       &       13,858      &    0 &   3,858      &     \checkmark      &      \checkmark     \\ \midrule 
        \textbf{\benchmarkName\ (ours)} &      \makecell{Block, C4,\\ Class, Flowchart,\\ Graph, Packet,\\Sequence, State} &   \imageCount       &     \imageCountReal   & 0   &\checkmark & \checkmark \\
        \bottomrule
    \end{tabular}
}
%\end{adjustbox}
    \vspace{-3mm}
    \caption{Comparison among different available corpora for technical diagram understanding}
    \label{tab:dataset-comparison}
    %\vspace{-5mm}
\end{table*}

\begin{itemize}[nosep,noitemsep]%,leftmargin=*]
\item We propose a large (\imageCount\ images) corpus (referred as \benchmarkName) of synthetically generated diverse technical diagrams, covering 8 different diagram types along with a manually created corpus of \imageCountReal\ hand-drawn technical diagrams of various types with the help of human participants. The corpus consists of triplets: synthetic technical diagram image, corresponding Mermaid\footnote{\url{https://mermaid.js.org/}} code, and corresponding image description. Collectively, the corpus spans a broad spectrum of commonly used diagram types, including flowcharts, block diagrams, state diagrams, graphs, C4 models, sequence diagrams, packet diagrams, and class diagrams. Fig. \ref{fig:entire-pipline} shows the entire pipeline.
\item To enable technical diagram understanding in models, we introduce various primary tasks (Image2code, Image2Description, Description2Code, and ImageEnhancement-via-Prompt) and self-supervision based tasks (CodeCompletion, Image-Code similarity). 
\item We experiment with a battery of pre-trained VLMs on these tasks via a zero-shot setup. Given the poor performance on zero-shot setting, we fine-tune Llama 3.2 11B-instruct \cite{Llama3.2-Vision(11B)} model on synthetic images on these tasks and obtain \modelName, which significantly improves the \textbf{ROUGE-L performance by 2.14x} and achieves \textbf{best all-round performance across all baseline models}. Further, we assess model generalization by manually evaluating several models on the Image2Code task with human participants. On real-world images, human evaluation reveals that we achieve the \textbf{minimum compilation errors} and average \textbf{F1 score} on correctly generating diagram structures by \textbf{6.97x}.
\item We create Github (\url{https://github.com/Exploration-Lab/TechING}) for the project having the details of dataset (\url{https://huggingface.co/datasets/Exploration-Lab/TechING}) and models (\url{https://huggingface.co/Exploration-Lab/LLama-VL-TUG}) for the research community.

\end{itemize}

\noindent The core contribution of this work lies in the integration and large-scale construction of a corpus that enables effective vision–language learning for hand-drawn technical diagrams and subsequent evaluation of existing VLMs on tasks related to the corpus. 

%\vspace{-2mm}
\section{Related Works} \label{sec:related}
%\vspace{-2mm}

Vision Language Models (VLMs) (e.g., PaliGemma-2 \citep{beyer2024paligemmaversatile3bvlm}, GPT-4o \citep{openai_gpt4o}, QwenVL \citep{bai2023qwenvlversatilevisionlanguagemodel}, DeepSeek-VL \citep{lu2024deepseekvlrealworldvisionlanguageunderstanding}, and BLIP-2 \citep{li2023blip2bootstrappinglanguageimagepretraining}) have been developed in recent years to process and understand image and text modalities together. VLMs have shown remarkable performance on a variety of tasks such as visual question answering \citep{dos2023visual,goyal2017makingvvqamatter,hudson2019gqanewdatasetrealworld,schwenk2022aokvqabenchmarkvisualquestion}, image captioning \citep{chen2015microsoftcococaptionsdata,young2014flickr30k,Agrawal_2019,chen2024compcapimprovingmultimodallarge,chen2015microsoftcococaptionsdata,young2014flickr30k,Agrawal_2019}, image-text retrieval \citep{zhang2024texttoimagediffusionmodelsgenerative}, text to image generation  \citep{li2024textguidedimageediting}, text-guided image editing \citep{Kembhavi2016ADI}, visual story telling \citep{zellers2019recognitioncognitionvisualcommonsense}, and document understanding \cite{li2024enhancing,aggarwal2023dublin}. However, technical diagram understanding is under-explored, with few recent works such as diagram understanding and reasoning \citep{Kembhavi2016ADI}, diagram summarization \citep{bhushan-lee-2022-block,bhushan-etal-2024-unveiling}, and question answering (\citet{masry2022chartqabenchmarkquestionanswering,methani2020plotqareasoningscientificplots,kahou2018figureqaannotatedfiguredataset}). Accordingly, various corpora have been introduced. Table \ref{tab:dataset-comparison} summarizes various available corpora along with \benchmarkName. As can be seen, \benchmarkName\ covers more diagram types than existing works and includes a much larger set of hand-drawn real-world images. In contrast, most existing real-world diagram corpora are scraped from structured sources, such as documentation or websites, where the diagrams are already clean and well-formatted. This makes \benchmarkName\ more challenging and realistic, as hand-drawn diagrams often exhibit variations in layout, noise, handwriting, and structure, reflecting practical scenarios more closely. %Prior work has introduced technical images, however, these are usually limited to one or two diagram types, and hence models trained on these fail to generalize to other diagram types. For instance, 
FlowLearn \cite{pan2024flowlearnevaluatinglargevisionlanguage} solely focuses on flowchart diagrams and lacks coverage of other diagram types, limiting its generalization scope. The diagram types we cover go beyond graph structures (i.e., some of these types cannot be represented as a graph, e.g., Packet, Sequence, C4, and Class Diagrams). Due to space limitations, more details are provided in App. \ref{app:sec-related-work}. %we provide details of other corpora and models in App. \ref{app:sec-related-work}. 

\vspace{-2mm}
\section{Corpus Creation and Tasks} \label{sec:corpus}                      \vspace{-2mm}

\begin{table*}[t]
    \centering
    \small                    
  
    \renewcommand{\arraystretch}{1.2}
    \begin{adjustbox}{width=1.0\linewidth}
    \begin{tabular}{ccccccccccc}
        \toprule
        \textbf{Dataset} & \textbf{Difficulty} & \textbf{Block} & \textbf{C4}  & \textbf{Class} & \textbf{Flowchart} & \textbf{Graph} & \textbf{Packet} & \textbf{Sequence} & \textbf{State} & \textbf{Total} \\
        \midrule

            \multirow{3}{*}{\textbf{D1}}
            & Easy     & 4402 (2.0) & 4738 (1.5)  & 6000 (2.0) & 1500 (4.5) & 3695 (3.3) & 1500 (3.0) & 6000 (2.3) & 5000 (3.4)& \textbf{32835} \\
            & Medium   & 2000 (4.5) & 3000 (4.5)  & 2000 (4.2) & 3000 (6.0)& 5001 (4.1) & 2981 (5.0) & 1500 (3.9) & 2500 (10.6) & \textbf{21554} \\
            & Hard     & 2500 (6.1) & 2500 (4.5) & 2500 (6.0) & 3500 (5.9)& 1729 (4.7) & 3500 (8.2) & 1786 (7.6) & 2000 (8.3) & \textbf{20443} \\

        \hline
            \textbf{D2} & Combined   & 2731 & 0  & 7646 & 0 & 5038 & 6683 & 8786 & 8753 & \textbf{39637} \\
            % &      &  &  &  &  &  &  &  & \textbf{Total} & \textbf{7454}\\
        \hline
\
            \textbf{D3} & Combined   & 74 & 62  & 74 & 72 & 77 & 53 & 58 & 75 & \textbf{545} \\

        \bottomrule
    \end{tabular}
    \end{adjustbox}
    \vspace{-2mm}
    \caption{\benchmarkName\ distribution across different tasks and diagram types for synthetic images and real-world hand-drawn images. The average number of primary components indicating the difficulty level is indicated in parentheses.}
    \label{tab:dataset_stats_transposed}
    %\vspace{-5mm}
\end{table*}

Due to the unavailability of large-scale real-world open-source hand-drawn images with their corresponding code, we programmatically generated a large number of synthetic images with different diagram types (Table \ref{tab:dataset_stats_transposed} and see App. \ref{app:sec:image-examples} for examples). These images are used to create the corpus \benchmarkName. In total, we constructed \imageCount\ samples comprising of synthetic Image–Mermaid Code–Description triplets and the derived image enhancement corpus for eight different diagram types; each diagram type has three difficulty levels (easy, medium, and hard) based on the number of components and relationship between them. To test performance in the real-world setting, we manually created a real-world images corpus for evaluation, having \imageCountReal\ hand-drawn technical images (on white board and paper) along with corresponding Mermaid code. %A detailed breakdown of the \benchmarkName\ corpus can be found in Table~\ref{tab:dataset_stats_transposed}.  Distribution of Evaluation set is described in \S\ref{sec:experiments}.

\vspace{-3mm}
\subsection{\textbf{\benchmarkName\ Creation}} 
\vspace{-1mm}

The corpus contains three parts as described below:

\noindent\textbf{{Image-Mermaid Code-Description Corpus (D1)}:} We programmatically created synthetic images and their description via Mermaid programming language (also see App. \ref{app:teching-creation}). As the first step, in order to populate  content in various components of a diagram, for each diagram type, we constructed discipline-specific (e.g., engineering, business, health) lists of keywords using GPT-4o. The lists were thoroughly reviewed to prevent harmful/toxic keywords. For instance, a keyword such as ``electrical'' could be assigned as a block name in a block diagram. The synthetic corpus creation system works as follows: based on the difficulty level (for instance, Easy (2-3 Headers), Medium (3-8 Headers), or Hard (6-9 Headers) in Packet diagrams -- Diagram of data packet used in communication where headers has information like IP Addresses, Flags etc.) and diagram type, the system first randomly selects a few components and edges that should be in the diagram, and subsequently selects a discipline and samples corresponding keywords. Similarly, according to the number of edges sampled, a pair of random blocks is connected with/without an edge label. Random sampling at each step was done using a uniform probability distribution. These were then populated into templatized Mermaid code, from which diagrams are rendered. We also manually created templates for generating descriptions corresponding to each diagram type. Mermaid code for each diagram in \textbf{D1} is parsed with regular-expression extractors to identify components, like blocks and edges, which were then inserted into the description templates to produce image–description pairs. The templates are simple python f-strings which are populated by the sampled keywords and edge information  updated in a loop line by line. Owing to the diversity of diagram structures, for each diagram type, a dedicated Python script is developed to automatically generate the corpus (see App. \ref{app:sec-dataset-stats} for corpus statistics). This pipeline mitigates the model’s inherent bias toward specific semantic content while strengthening its ability to learn diagram structures, thereby enabling robust training without requiring domain expertise. Also, all automatically generated data is created using predefined templates, ensuring consistency and correctness by design. As a result, no additional data cleaning or selection is required after generation. Note that the difficulty level of diagrams (based on headers) was decided empirically based on initial experiments where we found that basic building blocks (very easy examples) of a diagram were required to make the model learn the structure of the diagrams. For instance, in the case of Block Diagram, two nodes connected with an edge can be defined as a basic building block. We have found out that if these diagrams are not present in the training set the model struggles to learn to produce correct Mermaid codes. Hence, these building blocks are assigned as Easy level and subsequently higher levels are defined relative to this level.

\noindent\textbf{Why Mermaid?} Mermaid programming language, used for representing technical images, was selected due to its human readability, being editable, wide coverage across diagram types, and widespread popularity. Additionally, Mermaid also provides its CLI tools (compiler) to generate images through the command line interface, hence it can be easily integrated with python scripts for automatic creation of large scale corpus. It is also possible to use multiple diagramming languages, such as GraphViz\footnote{\url{https://graphviz.org/}}, to increase corpus size and complexity; however, our primary focus is not on training VLMs to learn multiple programming languages, but rather on enabling the model to understand and reason over technical images. Moreover, introducing multiple languages could increase variability in syntax, which may actually degrade fine-tuning performance by adding unnecessary complexity.

\noindent\textbf{{Image-Enhancement Corpus (D2)}:} Using \textbf{D1}, we created Image-Enhacement Corpus (\textbf{D2}) that consists of an incomplete image and a corresponding natural language prompt that directs a model to complete the image. For creating \textbf{D2}, we remove specific parts of technical images in \textbf{D1} by deleting triplets (two components connected by an edge in the Mermaid Code). The remaining code is then compiled again to get the incomplete image, which is coupled with an enhancement prompt. The prompt is generated automatically by verbalizing the removed triplet into textual description using Gemma3 (4B) model \cite{gemma3} resulting in Image-Enhancement corpus with images and enhancement prompts. The full dataset construction pipelines are illustrated in App. \ref{app:teching-creation}.
 
\noindent\textbf{{Real World Images Corpus (D3)}:} To test a model's technical image understanding and generalization capabilities, we created a corpus of hand-drawn images (with the help of humans) reflective of real-world setting. We curated \imageCountReal\ images with corresponding Mermaid Code, covering all diagram types. To ensure the Mermaid code accurately reflects the original image, it was compiled using the Mermaid compiler, and the generated image was compared to the hand-drawn image. These diagrams were hand drawn by annotators on different mediums (whiteboard, paper, and electronic whiteboard) to provide diversity in the corpus, which were subsequently digitized using a mobile phone (see examples in App. \ref{app:sec:image-examples}). The size of this corpus is relatively small compared to the synthetic corpus; this is primarily due to the significant difficulty in creating such a corpus. Each example requires not only manual drawing but also precise transcription into correct Mermaid code, which is highly time-consuming due to multiple iterations (compiling and comparing with the original image), prone to errors due to oversight, and hence difficult to scale. Even when using samples from existing hand-drawn corpora like books and research papers, generating the corresponding code remains a tedious and labor-intensive task. This corpus provides a realistic setting where images may be blurred or may have glare on them. The annotators were mainly graduate and under-graduate students (8 in total); they had knowledge of various diagram types, and they were not part of the project, hence minimizing potential biases. They were instructed to draw these diagram types based on how these diagrams look in real world scenarios. They performed the task on a pro-bono basis in order to contribute towards the advancement of technology. %App. \ref{app:sec:image-examples} provides some examples (D1, D2 and D3) from \benchmarkName. 

\vspace{-2mm}
\subsection{\noindent\textbf{Technical Diagram Understanding Tasks}}
\vspace{-1mm}

We selected many tasks inspired from real-life use cases to help the model learn alignment between image, code, and textual description. The tasks are categorized into primary (used for both training and evaluation) (App Fig. \ref{fig:tasks-pipeline}) and self-supervision-based (employed exclusively during training). See App. \ref{app:sec-tasks-examples} for example figures of all tasks. 

\noindent\textbf{{Image2Code (Primary)}:} Accurately translating images into structured code is essential for enabling automated diagram understanding and downstream reasoning. To this end, we frame the Image2Code task, where the model is required to generate corresponding Mermaid code for a given image. Once well-formed code is produced, rendering the technical diagram becomes straightforward using the Mermaid compiler.
  
% \vspace{-2mm}
\noindent\textbf{{Description2Code (Primary)}:} Enabling the translation of structural descriptions into executable code is crucial for rapid prototyping, documentation, and iterative editing of technical diagrams. We therefore frame the task of converting natural language descriptions -- capturing topological information such as the number of components and their connections -- into code, using code as an intermediate representation. This approach allows diagrams to be generated and refined efficiently through iterative prompts.

% \vspace{-2mm}
\noindent\textbf{{Image2Description (Primary)}:} Generating Descriptions from technical diagram images is essential for efficient search, retrieval, and question-answering tasks. These descriptions can also serve as contextual input for building advanced research agents, eliminating the need for the original image in later stages. By providing structured representations, the model enables diagrams to be indexed for both keyword-based queries (e.g., ``find all diagrams with retry loops'') and vector-based retrieval to identify structurally similar diagrams.

% \vspace{-2mm}
\noindent\textbf{{Image Enhancement via Prompt (Primary)}:} Certain applications like versioning require us to update our existing diagrams, but these diagrams are in the form of an image and not directly editable. We can train a model so that it can directly generate the code of the updated image when given an image and a natural language enhancement prompt as input. This updated code can then be rendered into the desired image using the Mermaid compiler.

\noindent\textbf{{Self Supervision Tasks}:} We introduce a set of self-supervision tasks (only for training) to learn fine-grained alignment between images, Mermaid code, and textual descriptions. In the \textbf{image enhancement via description task}, the model receives a description of the target image and identifies which components are present and which need to be added in the code.  In \textbf{code enhancement via prompt}, the model updates given Mermaid code based on an enhancement prompt, while in \textbf{code enhancement via description}, it completes missing parts of the code from a description. The \textbf{positive/negative image-code pair Q\&A task} involves predicting whether a given image and code pair match, while the \textbf{partial match image-code pair Q\&A} task requires identifying partial matches between incomplete and complete image-code pairs to capture sub-part relationships.

\vspace{-3mm}
\section{Methodology and Experiments} \label{sec:experiments}
\vspace{-2mm}

\begin{figure*}[htbp]
    \centering
    \includegraphics[width=\textwidth]{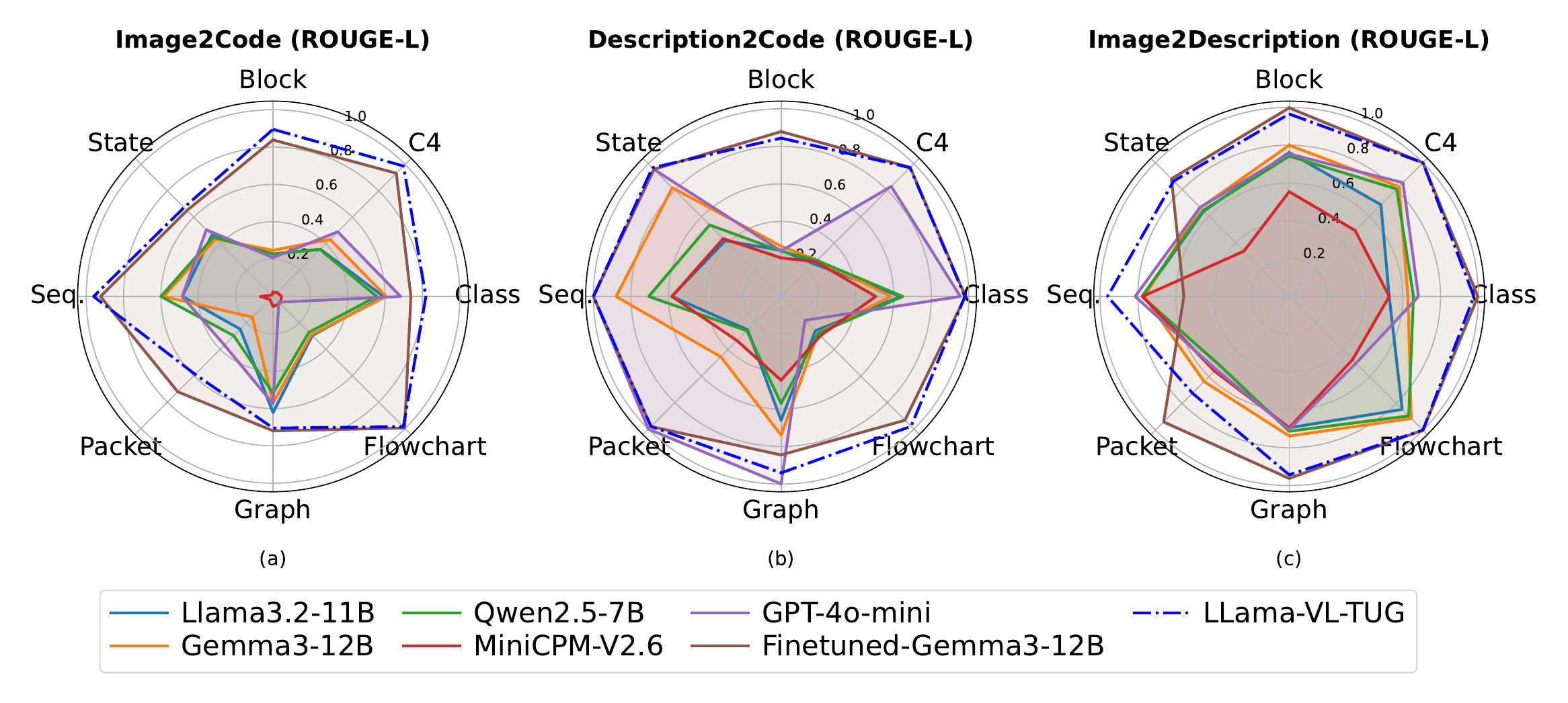}
    \vspace{-7mm}
    \caption{Results on Image2Code, Description2Code, and Image2Description tasks for the synthetic corpus (\textbf{D1}).}
    \label{fig:syn_radar}
    \vspace{-5mm}
\end{figure*}

We benchmark various VLMs/LLMs Llama3.2-11B-Vision-Instruct \cite{Llama3.2-Vision(11B)}, MiniCPM-V-2-6(8B) \cite{yao2024minicpmvgpt4vlevelmllm}, Qwen2.5-VL-7B-Instruct \cite{bai2025qwen25vltechnicalreport}, Gemma3-12B-Instruction-Tuned \cite{gemmateam2025gemma3technicalreport} and GPT-4o-mini \cite{openai_gpt4o_mini}) in zero-shot setting for the baseline (App. \ref{app:sec:baseline-models}). Zero-shot evaluation was deliberately chosen for the baselines to establish a fair and widely adopted point of comparison without requiring fine-tuning. Moreover, typically in a practical setting, VLMs/LLMs are used in a zero-shot setting. Our intent is to highlight the effectiveness of the proposed corpus and training pipeline, rather than to introduce new architectural or prompt-driven innovations. Given the poor performance of baselines (\S\ref{sec:results}), we fine-tuned a pre-trained Llama-3.2-11B-Vision-Instruct \cite{Llama3.2-Vision(11B)} model using LoRA \cite{lora} on the synthetic part of the corpus using tasks described earlier.  Subsequently, the trained model is tested on real-world data and evaluated by humans to gauge its practical applicability and generalizability. To further validate our corpus and fine-tuning strategy, we also fine-tuned Gemma3-12B-Instruction-Tuned and evaluated it on the Synthetic Corpus Evaluation set across the three primary tasks, following the same procedure as with our fine-tuned model, \modelName. The results (Fig. \ref{fig:syn_radar}) are consistent with those of \modelName, supporting the effectiveness of our corpus and fine-tuning approach.
\begin{table*}[ht]
    \centering
    % {\fontsize{7}{9}\selectfont
    \renewcommand{\arraystretch}{1.2}
    \begin{adjustbox}{width=1\textwidth}
    \begin{tabular}{lcllllcllll}
        \toprule
        \textbf{Model} & \textbf{Diag}  & \textbf{CEr$^\star$$\downarrow$} & \textbf{CBl$^\star$$\uparrow$} & \textbf{CEd$^\star$$\uparrow$}  & \textbf{CLE$^\star$$\uparrow$/CAM$^\star$$\uparrow$}  &  \textbf{Diag} & \textbf{CEr$^\star$$\downarrow$} & \textbf{CBl$^\star$$\uparrow$} & \textbf{CEd$^\star$$\uparrow$}  & \textbf{CLE$^\star$$\uparrow$/CBi$^\star$$\uparrow$}   \\
        \midrule        
        
        \textbf{Llama3.2-11B} & \multirow{6}{*}{\rotatebox{90}{\textbf{Block}}} & 0.76 & 0.33 &	0.20	& 0.05  &  \multirow{6}{*}{\rotatebox{90}{\textbf{Graph}}} & 0.56  & 0.52 &	0.37 &	0.03       \\
        \textbf{Llama-VL-TUG (ours)} & &	\textbf{0.03} (25x) & 0.80 (2.4x)	& 0.58 (2.9x) &	0.35 (7x)   & & \textbf{0.00} (\textasciitilde2x) & \textbf{0.96} (1.8x) &	\textbf{0.88} (2.4x)&	\textbf{0.78} (26x)     \\
        \textbf{Gemma3-12B} & & 0.08 & \textbf{0.92}	& \textbf{0.77} &	0.38  & &	0.13 & 0.86 &	0.71 &	0.35     \\
        \textbf{Qwen2.5-VL-7B} &  &	0.27  & 0.81 &	0.66 &	0.41  & &	0.23  & 0.80 &	0.72 &	0.36    \\
        \textbf{MiniCPM-V-2-6} & &	0.16 & 0.81	& 0.50 &	0.19  & & 0.30 & 0.69 &	0.59 &	0.08    \\
        \textbf{GPT-4o-mini} & &	0.07 & \textbf{0.92} &	0.60 &	\textbf{0.43}  & &	0.09 & 0.92 &	0.83 &	0.75      \\
         \midrule
         
        \textbf{Llama3.2-11Bt} &	\multirow{6}{*}{\rotatebox{90}{\textbf{C4}}} & 0.06 & 0.07 &	0.20 & NA   & \multirow{6}{*}{\rotatebox{90}{\textbf{Packet}}} &	0.66  & 0.00 & NA  &	0.00    \\
        \textbf{Llama-VL-TUG (ours)} & &	\textbf{0.03} (2x) & \textbf{0.71} (10x)&	0.35 (1.8x)& NA   & & \textbf{0.04} (16.5x)  & \textbf{0.40} (\textasciitilde40x)	& NA  & \textbf{0.15} (\textasciitilde15x)   \\
        \textbf{Gemma3-12B} &	& 0.55  &0.51 &	0.21 & NA   & & 0.23 & 0.00	& NA & 0.00	    \\
        \textbf{Qwen2.5-VL-7B} & &	0.68 & 0.36 &	0.18 & NA  & & 0.51  & 0.00	& NA   & 0.00    \\
        \textbf{MiniCPM-V-2-6} & &	0.76  & 0.18 &	0.01  & NA   & & 0.51 & 0.00	& NA  & 0.00  \\
        \textbf{GPT-4o-mini} & &	0.37 & 0.64 &	\textbf{0.41} & NA  & & 0.25 & 0.00	& NA  & 0.00 \\
         \midrule
        
        \textbf{Llama3.2-11B} &	\multirow{6}{*}{\rotatebox{90}{\textbf{Class}}} & 0.57 & 0.60 & 	0.36  &	0.56   & \multirow{6}{*}{\rotatebox{90}{\textbf{Sequence}}} &	\textbf{0.00} & 0.59 & NA &	0.34   \\
        \textbf{Llama-VL-TUG (ours)} & & \textbf{0.03} (19x) & 0.79 (1.3x) & 0.66 (1.8x)  &	0.71 (1.3x)  & &	\textbf{0.00} (1x) & 0.78 (1.3x)& NA &	0.39 (1.1x)       \\
        \textbf{Gemma3-12B} & & 0.86 & 0.86 & \textbf{0.86}  &	0.86  &	& 0.10  & \textbf{0.93}	& NA & 0.64   \\
        \textbf{Qwen2.5-VL-7B} & &	0.32  & 0.72 & 0.45  &	0.70  & &	\textbf{0.00} & 0.67 & NA &	0.45     \\
        \textbf{MiniCPM-V-2-6} & &	0.92 & 0.09 &	0.02  &	0.00  & &	0.57 & 0.54 & NA &	0.35     \\
        \textbf{GPT-4o-mini} & &	\textbf{0.03}  & \textbf{0.98} &	\textbf{0.86}  &	\textbf{0.93}  & & 0.02 & 0.90 & NA &	\textbf{0.72}   \\
         \midrule
        
        \textbf{Llama3.2-11B} &	\multirow{6}{*}{\rotatebox{90}{\textbf{Flowchart}}} & 0.86  & 0.25 &	0.21 &	0.18   & \multirow{6}{*}{\rotatebox{90}{\textbf{State}}} &	0.73 & 0.34 &	0.19 &	0.00     \\
        \textbf{Llama-VL-TUG (ours)} & &	\textbf{0.00} (\textasciitilde6x) & 0.86 (3.4x) &	0.63 (3x) &	0.29 (1.6x)   & &	0.16 (4.6x)  & 0.52 (1.5x) &	0.31 (1.6x) &	0.19 (\textasciitilde19x)       \\
        \textbf{Gemma3-12B} &	& \textbf{0.00}  & \textbf{0.97} &	\textbf{0.88} &	0.81    & & \textbf{0.03} & 0.76 & 0.61 &	0.52	  \\
        \textbf{Qwen2.5-VL-7B} & &	0.33  & 0.74 &	0.70 &	0.46   & & 0.44 & 0.52 &	0.38 &	0.34      \\
        \textbf{MiniCPM-V-2-6} & &	0.31 & 0.69 &	0.44 &	0.11   & &	0.24 & 0.68 &	0.29 &	0.05     \\
        \textbf{GPT-4o-mini} & &	0.04  & 0.95 &	\textbf{0.88} &	\textbf{0.85}   & & \textbf{0.03} & \textbf{0.81} &	\textbf{0.65} &	\textbf{0.56}    \\
         % \midrule

        \bottomrule          
    \end{tabular}
    \end{adjustbox}
% }
    %\vspace{-2mm}
    \caption{Human evaluation scores for Image2Code task on the Real-world images corpus \textbf{D3}. Improvement factor of \modelName\ after fine-tuning over the vanilla variant is indicated in parentheses. Except for Compilation Error \textbf{CEr}, all other columns show F1 score. $^\star$ \textbf{CEr:} Compilation Error; \textbf{CBl:} Correct Blocks (blocks/nodes/classes/headers); \textbf{CEd:} Correct Edges; \textbf{CLE:} Correct Labeled Edges; \textbf{CAM:} Correct Attributes \& Methods (for Class Diagrams); \textbf{CBi:} Correct Bits (for Packet Diagrams); NA: Not Applicable}
    \label{tab:image_2_code_real}
    %\vspace{-5mm}
\end{table*}

\noindent{\textbf{Fine-tuning:}}
We fine-tuned Llama3.2-11B-Vision-Instruct using LoRA (image encoder as well as text decoder) on the combination of all tasks (both Primary and Self Supervision) using \textbf{D1} and \textbf{D2}. While there are several options available with regard to the choice of VLM, due to limited compute availability, we focused on fine-tuning just two VLMs. Out aim is to validate the effectiveness of the training pipeline and code-generation approach, and insights gained from these experiments can easily be generalized to other models. Our goal is to demonstrate the benefits of the methodology rather than compare fine-tuning performance across every model. 
%\TN{We fine-tuned only the LLaMA model to validate the effectiveness of our training pipeline and code-generation approach, since fine-tuning all baseline models would be computationally prohibitive and unlikely to change the insights, as our goal is to demonstrate the benefits of the methodology rather than compare fine-tuned performance across every model.}
During fine-tuning, a task is selected randomly (all tasks having equal probability), and accordingly, a data point is also sampled randomly with suitable input-output pair selected from the corpus, which then goes into the model, and gradient updates are made (see App. \ref{app:sec:hyperparam} for hyperparameter details). For instance, in Image2Code task, a sample from \textbf{D1} corpus with Image as input and Mermaid Code as target output is selected for the forward and backward pass, respectively. Eventually, the model is fine-tuned on a dynamic mixture of tasks sampled in a random order at each training step. This helps the model generalize across tasks, handle different input types, and transfer knowledge to various technical reasoning scenarios. We refer to the model as \modelName\ (LLama-Vision Language-Technical image Understanding and Generation). To improve the diversity of the data and bring it closer to a real-world setting, we also performed data augmentation on every training sample by performing various transformations while training, such as blur, noise, lighting changes, and distortions (App. \ref{app:sec-augmentations}).

\noindent{\textbf{Evaluation:}}
We evaluated all the models on synthetic  as well as real-world diagrams (Fig. \ref{fig:entire-pipline}). The synthetic evaluation corpus was created by the same strategy as \textbf{D1}, i.e., for each diagram type, we programmatically generated $500$ Image-Mermaid Code-Description triplets and selected the input output pair according to the task. For the real world evaluation, real world corpus \textbf{D3} was used. For the synthetic diagrams, we conducted evaluation of our fine-tuned model and the baselines on the three primary tasks, Image2Code, Description2Code, and Image2Description using standard metrics, BLEU \cite{papineni-etal-2002-bleu}, SACREBLEU \cite{post-2018-call}, METEOR \cite{banerjee-lavie-2005-meteor}, chrF \cite{popovic-2015-chrf}, BLEURT \cite{sellam-etal-2020-bleurt}, and ROUGE-L \cite{lin-2004-rouge} (deatils in App. \ref{app:sec-metrics}). Although order-agnostic evaluation tools such as \textit{mtool} \citep{oepen-etal-2019-mrp} are available, they are limited to graph-based representations; therefore, we restrict our evaluation to the aforementioned automatic metrics for consistency across graph-based and non-graph based diagrams. The real-world images corpus (\textbf{D3}) was used to evaluate on Image2Code task with the same aforementioned metrics. All the experiments were conducted in a zero shot setting, and seeds were saved for reproducibility. 

\noindent{\textbf{Human Evaluation:}} 
Automatic evaluation metrics such as BLEURT and ROUGE-L, while widely used in natural language tasks, are not well-suited for evaluating the Image2Code task. These metrics primarily capture surface-level textual similarity and struggle to account for structural correctness or semantic equivalence in code, where multiple variations may yield the same diagram (order-agnostic). Owing to these limitations, and to ensure reliable assessment, we conducted human evaluation on \textbf{D3}, focusing on issues such as incorrect diagram types, compilation errors (errors in the Mermaid code syntax that cause image generation to fail), spurious or missing structures (blocks, edges, attributes, etc.). Eight human annotators (participating on a pro bono basis) (different from those who created diagrams) were asked to compare the original image with the image regenerated from the code (for the hand-drawn image) generated by the model, and they were asked to evaluate on various metrics. For the Image2Code task, the evaluators first evaluate \% of compilation errors of the generated Mermaid code. Among the successfully compiled images, the evaluators compute models' precision, recall, and F1 score for each structure (blocks, edges, attributes, etc.), taking the hand-drawn image as the ground truth (details in App. \ref{app:sec-metrics}). 

\noindent{\textbf{Ablation Study:}} 
An ablation study was also performed where we fine-tuned the model only on primary tasks (without self-supervision tasks) and evaluated on the Image2Code task using \textbf{D3} to compare its performance against \modelName.

%\vspace{-1.7mm}
\section{Results, Analysis and Discussion} \label{sec:results}
%\vspace{-1.9mm}

\noindent\textbf{Automatic Eval. Results and Analysis:} Fig. \ref{fig:syn_radar} shows radar plots (with ROUGE-L scores per diagram type) for Image2Code, Description2Code, and Image2Description. In \textbf{Image2Code} task \modelName\ consistently outperforms all baselines, including GPT-4o-mini on all 6 metrics (results in App. Table \ref{tab:image_2_code_metrics}). \modelName\ has near perfect ROUGE-L scores on C4, Flowchart and Sequence Diagrams (Fig. \ref{fig:syn_radar}(a)). The effectiveness of finetuning on \benchmarkName\ is evident, as our model improves the average ROUGE-L scores (average over all diagram types) of Llama3.2-11B-Vision-Instruct by 2.29x and over GPT-4o-mini by 4.53x. A huge improvement (22.6x) is seen on the flowchart type image over GPT-4o-mini. In \textbf{Description2Code} task (Fig. \ref{fig:syn_radar}(b)) highlights that \modelName\  has the best average ROUGE-L performance. Our model is competitive with GPT-4o-mini on Graph and Packet diagrams and outperforms all other baselines by a significant margin on all other diagrams. Similar results are obtained on the remaining 5 metrics: BLEU, SACREBLEU, METEOR, chrF, and BLEURT  (see App. Table \ref{tab:summary_2_code_metrics}). The fine-tuned model improves the ROUGE-L scores of Llama3.2-11B-Vision-Instruct by 2.75x and over GPT-4o-mini by 1.88x on this task. Major improvements are observed in Block (3.48x) and Flowchart (5.45x) over GPT-4o-mini. Similar results are observed in the \textbf{Image2Description} task. Fig~\ref{fig:syn_radar}(c) shows that \modelName\  has the best all-round ROUGE-L performance. The model  outperforms GPT-4o-mini and all other baselines by a significant margin on State, Sequence, Graph, Class, C4, Block and Flowchart diagrams and is competitive on Packet diagrams (more results in App. Table \ref{tab:topological_summary_metrics}). The fine-tuned model improves the average ROUGE-L scores of Llama3.2-11B-Vision-Instruct by 1.37x and outperforms GPT-4o-mini by 1.38x. % on this task. 

\noindent\textbf{Human Evaluation:}
Table \ref{tab:image_2_code_real} summarizes the results of human evaluation for Image2Code performance on Block and C4 diagrams. We report F1 scores for evaluating model performance in generating correct structural components—blocks/nodes/classes/headers, edges, labeled edges, attributes and methods, and bits—from real-world hand-drawn images (detailed results in App. Table \ref{tab:auto_real_world}). For scoring block-type structures, both node detection and correct block name recognition are required. Edges are scored only when both the source and target nodes match (node-edge-node triplet), and labeled edges additionally require the correct label. This strict scoring allows us to implicitly capture different types of errors, such as OCR/text recognition failures, node detection errors, and edge routing mistakes. 

Our findings show that \modelName\ achieves the lowest percentage of compilation errors, with a 9.8x reduction compared to Llama3.2-11B-Instruct. Moreover, fine-tuning on \benchmarkName\ significantly boosts the F1 scores over Llama3.2-11B-Instruct: 7.7x on blocks, 2.23x on edges, and 10.9x on labeled edges. None of the baseline models correctly identified the diagram type for Packet Diagrams. Since Packet Diagrams are structurally distinct from regular graph-type diagrams, we assigned zero scores for incorrect generations. Notably, only the fine-tuned model was able to accurately detect and generate Packet Diagrams. In graph diagrams, \modelName\  demonstrates strong performance, correctly generating all nodes, edges, and labeled edges. The evaluation also highlights strengths and weaknesses of baseline models: for instance, GPT-4o-mini performs well on blocks, edges, and attributes in class and state diagrams but shows mediocre performance on blocks and bit labels in packet diagrams, while Gemma3-12B reliably generates blocks across diagram types but struggles with accurate edge generation.
 
\begin{figure}[t]
    \centering
    \includegraphics[width=0.5\textwidth]{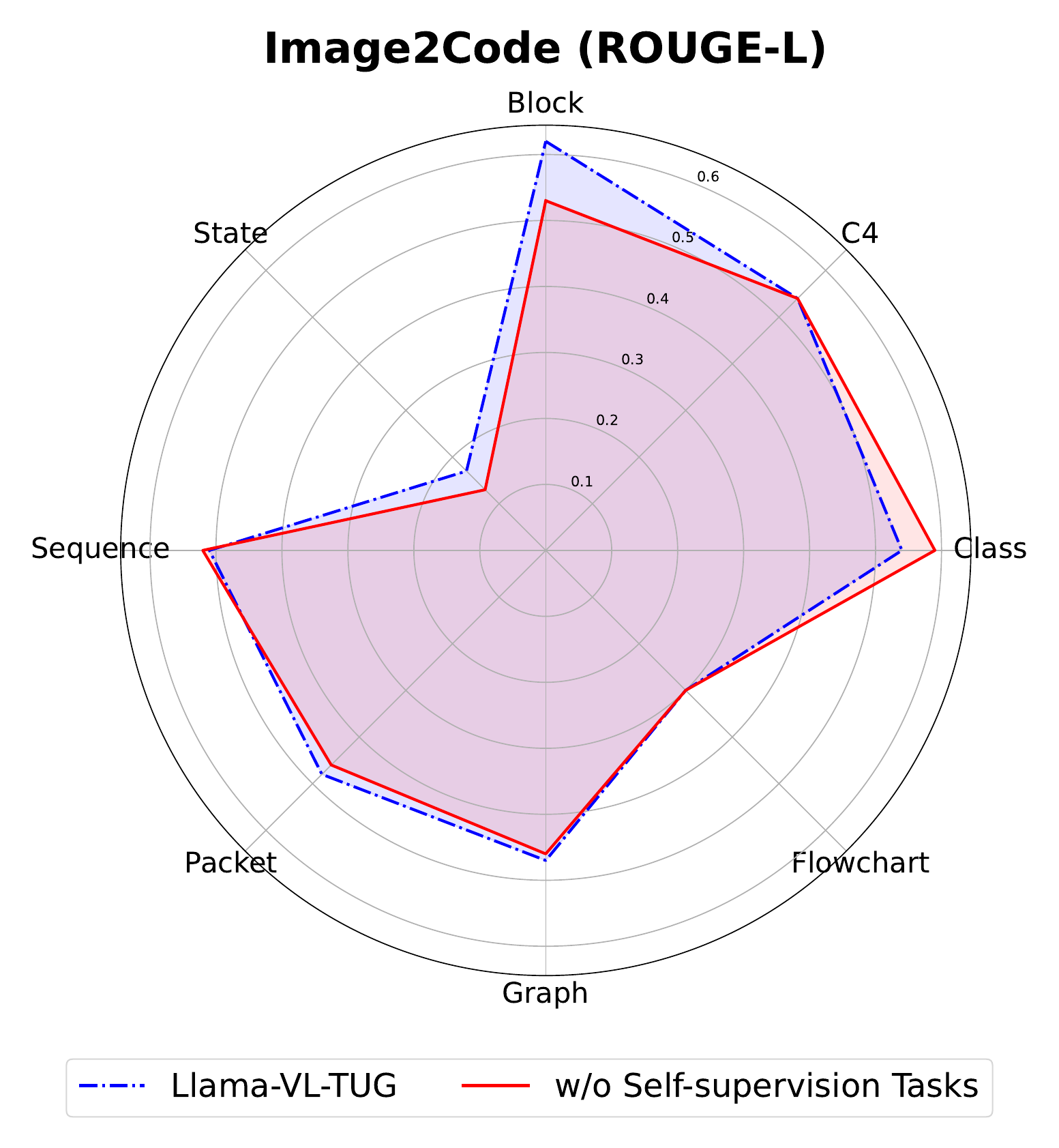}
    \vspace{-8mm}
    \caption{Ablation Study results (Real-World: \textbf{D3})}
    \vspace{-5mm}
    \label{fig:ablation-radar}
\end{figure}

\noindent\textbf{Ablation Study:} Role of self-supervision tasks in fine-tuning is shown in Fig. \ref{fig:ablation-radar}. \modelName\  is shown to outperform the ablated variant on real-world hand-drawn corpus \textbf{D3} in the Image2Code task across nearly all diagram types on ROUGE-L score. Most notably, our model  achieves the highest score on block and state diagrams, indicating that self-supervision aids in understanding structurally rich and semantically diverse visual elements. Interestingly, for some categories like sequence and graph, the gap between ablated models and \modelName\ is smaller, indicating that these tasks may be less dependent on the self-supervision signals but still benefit overall from the full training pipeline, as the representations learned via self-supervision can transfer across diagram types and help in enhancing performance.

\noindent An additional ablation direction would involve isolating the contribution of individual primary tasks. However, preliminary experiments with continual training—where the model was trained sequentially on Image2Code, Description2Code, and self-supervised image enhancement—revealed that performance on earlier tasks remained stable even after training on subsequent tasks. This suggested that the primary tasks cultivate overlapping capabilities, rendering task-specific ablations unlikely to yield meaningful insights. Consequently, we did not pursue this ablation study. More details and results in App. \ref{app:section-I}

\noindent\textbf{Discussion and Error Analysis:} 
 \noindent Real-world diagrams have noise and incompleteness, and a lot of variations in the user styles, leading to errors in the code generated by models. Resolving this can potentially require a huge amount of data for training. We have explored the use of image augmentation and found it to be useful for improving model performance. Although \modelName\ was trained using synthetic images in \benchmarkName\ only, it demonstrated good generalization over real-world images (also see Limitations). Further, we find that most of the existing VLMs can reliably generate only graph-like diagrams, while for other types, the Mermaid code often fails to compile and includes hallucinated elements. The performance on edges is notably weak, with labeled edges performing even worse, including in sequence diagrams where messages act as labeled edges. Models also frequently generate the wrong diagram type, most notably in the case of packet diagrams, where all models fail to detect or generate them. Notably, only after fine-tuning does our model successfully detect and generate packet diagrams. 

\vspace{-3mm}
\section{Conclusion} \label{sec:conclusion}
\vspace{-2mm}
We address a low-resource setting and introduce a large synthetic corpus \benchmarkName\ with different complexity levels (Easy, Medium, Hard) to enhance the technical image understanding of VLMs. We benchmark various open 10B parameter models and closed source GPT-4o-mini model. We showed that fine-tuning on a pretrained VLM (resulting in \modelName) using a synthetic corpus via various tasks helps to improve real- world hand-drawn technical diagram understanding. Both automated and human evaluations show the efficacy of our approach compared to existing baseline models. In the future, we plan to continue to grow the corpus of real-world hand-drawn images to capture more variations in data (also see limitations).

\section*{Limitations} \label{sec:limitations}
%\vspace{-3mm}

\noindent\textbf{Corpus Size:} In this paper, we made initial steps towards solving a hard problem due to large variability and domain specificity by introducing a large, diverse corpus. Since the synthetic images in \benchmarkName\  corpus were generated through a randomized strategy with few random templates, this may not capture all the possible variations that humans may introduce while creating diagrams by hand. To address these, we added \imageCountReal\ real-world hand-drawn diagrams to the corpus for evaluation. However, we anticipate encountering significantly greater variation in real-world scenarios. For example, the style of arrows, the way nodes and blocks are drawn, and even the diagram type vary widely in the real world. This may likely contribute to the decrease in performance, as the model is not specifically trained on all possible variations that are observed in the real-world images. Nevertheless, we plan to continue growing the corpus of real world hand-drawn images to capture more variations.  

\noindent It is also difficult to manually evaluate large diagrams generated by VLMs; thus, we restricted the human evaluation on images with not more than 10-15 structures/blocks. 

\noindent\textbf{Synthetic Corpus:} In this paper, we used a programmatic approach for generating synthetic images, as this guarantees consistency and correctness of technical diagrams. In our initial experiments, we tried diffusion and CLIP based models for technical image generation; however, these did not perform well and further research is required in training these models, which we plan to address in the future.

\noindent Real-world diagrams can have a lot of variations. For example, the style of arrows, the way nodes and blocks are drawn, and even the diagram type vary widely in the real world. In this work, we covered a subset of variations. Further work is needed to generate many variations of the same user-drawn image to increase the variety and create a large-size training corpus automatically. Given the challenges of manually evaluating large diagrams -- where models may alter the diagram type or introduce spurious content -- we deliberately limit human evaluation to diagrams with at most 10–15 structures or blocks. This controlled scope ensures consistency, reliability, and feasibility in the evaluation process.

\noindent Moreover, our dataset does not encompass the full range of highly irregular or informal hand-drawn sketches encountered in practice. While extremely messy diagrams remain challenging, the system generally produces a near-correct structural backbone that significantly reduces manual effort compared to diagram reconstruction from scratch. Future work should focus on improving robustness to more degraded visual inputs commonly found in engineering workflows.

\noindent \textbf{Training Strategy:} We experimented with a common finetuning strategy in a zero-shot setting for fine-tuning the VLM and did not explore complex techniques of context engineering, few-shot learning, etc., which we would like to explore in future work. We also acknowledge that fine-tuning all available open-source models could provide a more comprehensive comparison. However, due to computational constraints, we focused on fine-tuning a single model. Despite this limitation, our experiments demonstrate the effectiveness of the corpus through the strong performance of our finetuned model. Future work can explore fine-tuning additional models as more computational resources become available. Moreover, we would also like to explore other model architectures, such as mixture of experts VLMs. 

\noindent \textbf{Tasks:} Further, in this paper, we experimented with a few tasks only (e.g., Image2Code, Description2Code, and Code2Description). While there can be several other tasks that could be used for training, we focused on the ones that we found practically relevant. 

\noindent We deliberately adopt a simple, unified pipeline for diagram understanding to clearly evaluate the effectiveness of our corpus and training approach. This keeps the current study focused, while leaving systematic comparisons with other agentic pipelines like DiagramAgent \cite{11093195} and exploration of hybrid or tool-augmented methods for future work.

% \lipsum[2-4]

%\vspace{-2mm}
\section*{Ethical Considerations} \label{sec:ethical}
%\vspace{-3mm}
To the best of our knowledge, we do not see any direct societal harm from this work. In fact, this work aims to solve a problem that could benefit society at large.

% % Bibliography entries for the entire Anthology, followed by custom entries
% %\bibliography{anthology,custom}
% % Custom bibliography entries only
% \bibliography{custom}
\bibliography{references}

\clearpage
\newpage

\appendix

\appendix

\section*{Appendix}

%%%%%%%%%%%%%%%%%%%%%%%%%%%

\titlecontents{section}[18pt]{\vspace{0.05em}}{\contentslabel{1.5em}}{}
{\titlerule*[0.5pc]{.}\contentspage} % Set the formatting for appendix sections in the table of contents
% % for list of tables
% \titlecontents{table}[0pt]{\vspace{0.05em}}{\contentslabel{1em}}{}
% {\titlerule*[0.5pc]{.}\contentspage} % Set the formatting for appendix tables in the list of tables

% for list of figures
%\titlecontents{table}[0pt]{\vspace{0.05em}}{\contentslabel{1em}}{}
\titlecontents{table}[0pt]{}{\contentslabel{1em}}{}
{\titlerule*[0.5pc]{.}\contentspage} % Set the formatting for appendix tables in the list of tables

\startcontents[appendix] % Start the table of contents for the appendix
\section*{Table of Contents} % Title for the appendix table of contents
%\addcontentsline{toc}{section}{Table of Contents} % Add the appendix table of contents to the main table of contents
\printcontents[appendix]{section}{0}{\setcounter{tocdepth}{4}} % Print the table of contents for the appendix

\startlist[appendix]{lot} % Start the list of tables for the appendix
\section*{List of Tables} % Title for the appendix list of tables
%\addcontentsline{lot}{section}{List of Tables} % Add the appendix list of tables to the main list of tables
\printlist[appendix]{lot}{}{\setcounter{tocdepth}{1}} % Print the list of tables for the appendix

\startlist[appendix]{lof} % Start the list of tables for the appendix
\section*{List of Figures} % Title for the appendix list of tables
%\addcontentsline{lot}{section}{List of Tables} % Add the appendix list of tables to the main list of tables
\printlist[appendix]{lof}{}{\setcounter{tocdepth}{1}} % Print the list of tables for the appendix

\newpage

%%%%%%%%%%%%%%%%%%%%%%%%%%%

\section{Related work} \label{app:sec-related-work}

The domain of technical image understanding has recently started gaining attention, and researchers have focused on tasks like diagram understanding and reasoning (\citet{Kembhavi2016ADI}), diagram summarization \citet{bhushan-lee-2022-block,bhushan-etal-2024-unveiling}, and Question Answering (\citet{masry2022chartqabenchmarkquestionanswering,methani2020plotqareasoningscientificplots,kahou2018figureqaannotatedfiguredataset}). Accordingly, various datasets have been proposed. CBD \citep{bhushan-lee-2022-block} was created by web crawling from different search engines for the summarization task of block diagrams; it contained a total of 502 samples. FlowchartQA \citep{tannert-etal-2023-flowchartqa} comprised of 1M programmatically created flowchart images using graphviz\footnote{https://graphviz.org/} by selecting random names and edge labels. The dataset contained 6M questions on geometry and topology using a range of templates. BD-EnKo \citep{bhushan-etal-2024-unveiling}, a multilingual, mostly synthetic summarization dataset of 47k images of the English language (91 real world images) was generated by prompting GPT-3.5 to give Mermaid code of various types and random themes of diagrams. The summary was also generated by GPT-3.5 by providing the generated code as a prompt. FlowVQA \citep{singh-etal-2024-flowvqa}, a QA dataset, used real-world examples of WikiHow articles, Instructables DIY blogs, and FloCo \citep{10.1007/978-3-031-41734-4_31} to create structured summaries and then again converting the summary into Mermaid code by GPT-4. Questions were also generated by GPT-4 by giving a summary and Mermaid code and a template question. This dataset contained 2,272 images and 22,413 questions. \citet{ai-etal-2024-advancement} introduced a graph dataset with 3,929 English and 2,747 Chinese image-text pairs, where images were scraped from  search engines, and text in the form of questions with candidate answers were generated by GPT-4V. DoCo \cite{li2024enhancing} employed Contrastive Learning in the pretraining phase and then using the encoding in the finetuning phase of an LLM. CoG-DQA \cite{Wang_2024_CVPR} CoG-DQA leverages LLMs to guide diagram parsing tools (DPTs) through the guiding chains, enhancing the precision of diagram parsing for the QA task.

\noindent The FlowLearn dataset \cite{pan2024flowlearnevaluatinglargevisionlanguage} is designed to improve the understanding of flowcharts, with a particular emphasis on scientific contexts. It consists of two subsets: Scientific Flowcharts (3,858 diagrams) curated from real-world sources, and Simulated(through Python scripts) Flowcharts (10,000 diagrams) generated to augment coverage. While FlowLearn provides valuable resources for studying flowchart comprehension, it is limited in scope as it focuses exclusively on a single diagram type, namely flowcharts. Additionally, it contains no hand-drawn real-world images, relying only on clean, structured diagrams, which reduces its suitability for evaluation on real-world hand-drawn images. Another similar yet distinctive domain related to our work involves the use of SVG for image generation, as seen in approaches like \citet{rodriguez2024starvectorgeneratingscalablevector} and \citet{wu2024chat2svgvectorgraphicsgeneration}. While their method targets simple images and employs SVG code as an intermediate representation, our work focuses on the understanding of technical diagrams using Mermaid code. 

\noindent ChartReader \cite{Cheng_2023_ICCV} focuses on understanding numerical information from charts, rather than structural or diagrammatic content, and supports chart types such as bar, line, and pie charts in tasks like Chart-to-Table, ChartQA, and Chart Summarization. CHAIN-OF-REGION \cite{li2025chainofregion} improves VLM performance on scientific diagrams by decomposing diagrams into regions using computer vision techniques. In contrast, our work focuses on diagram regeneration and editing via an intermediate code representation for structural understanding across diverse diagram types. 

\noindent DiagramGenBenchmark and DiagramAgent \cite{11093195} introduce text-to-diagram generation, producing structured, editable diagrams from text, primarily covering flowcharts, model architecture diagrams, and mind maps, targeting a completely different set of diagrams from our work.

\noindent Table \ref{tab:dataset-comparison} summarizes various available datasets along with \benchmarkName. As can be seen, \benchmarkName\ covers more diagram types than existing works and includes a much larger set of hand-drawn real-world images. 
%%%%%%%%%%%%%%%%%%%%%%%%%%%%%

\section{\benchmarkName\ Image Examples} \label{app:sec:image-examples}

\begin{figure*}
    \centering
    \includegraphics[width=1.0\linewidth]{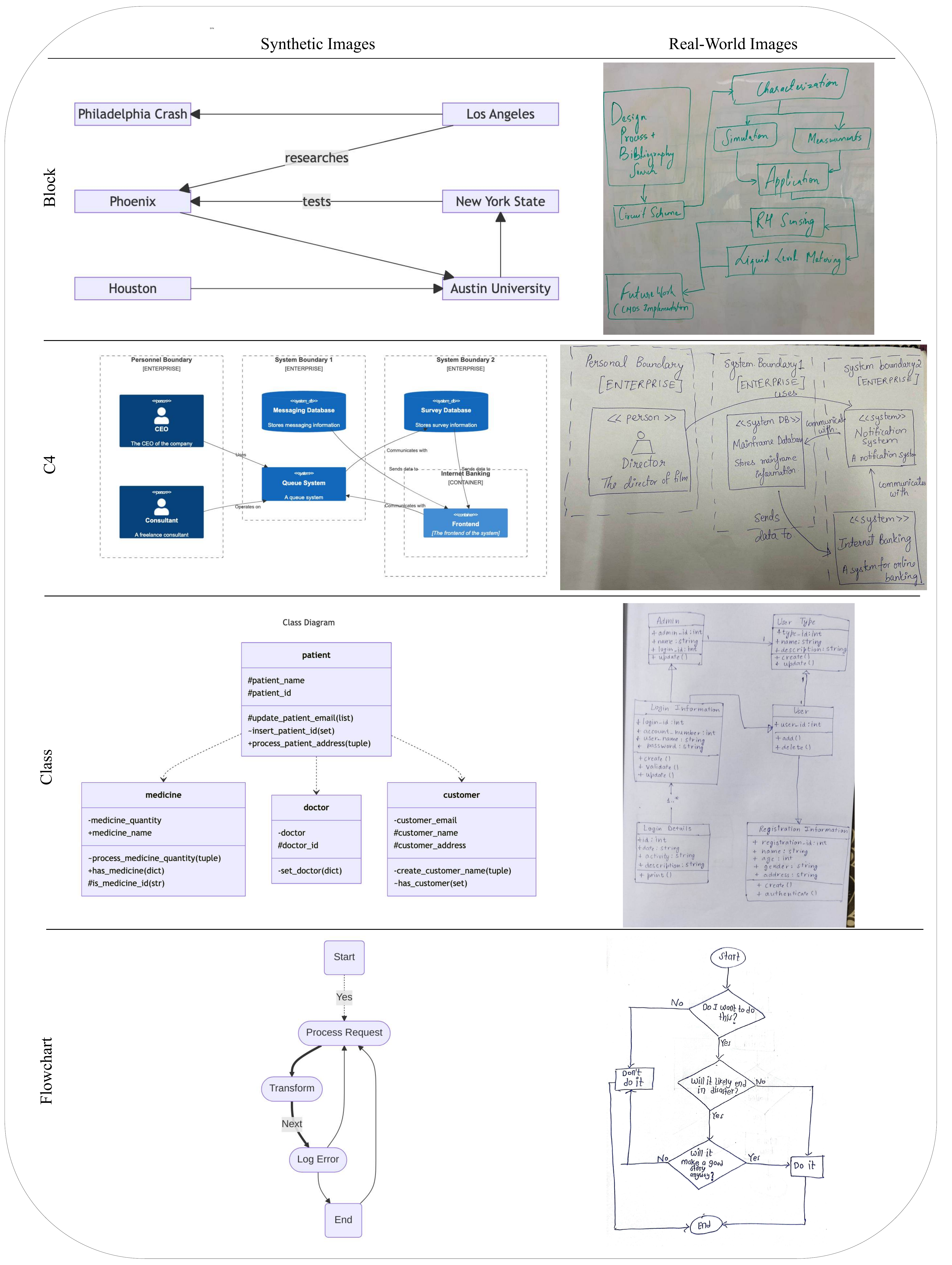}
    \caption{Synthetic and Real-World Examples for Block, C4, Class, Flowchart Diagrams}
    \label{fig:image_examples1}
\end{figure*}

\begin{figure*}
    \centering
    \includegraphics[width=1.0\linewidth]{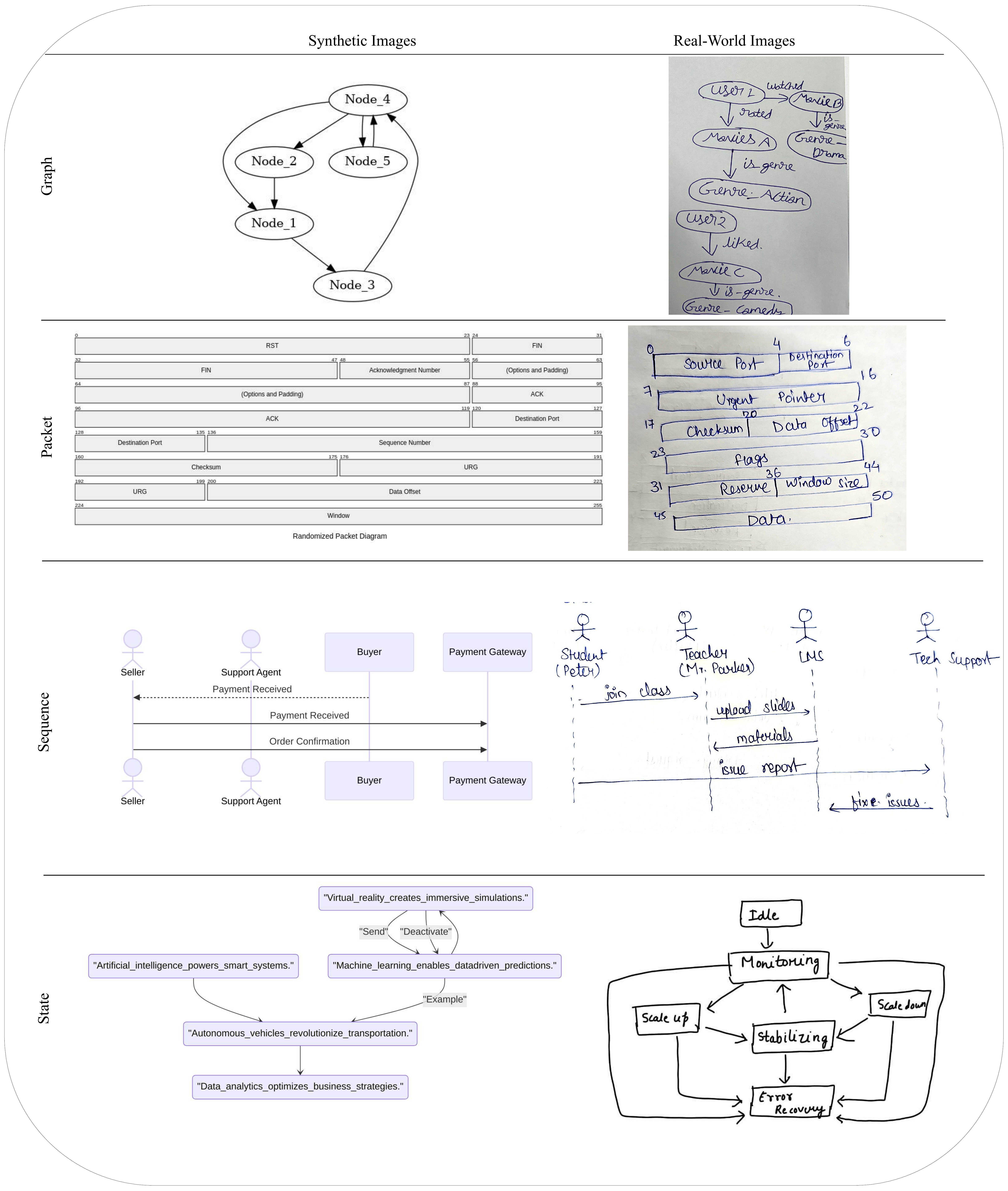}
    \caption{Synthetic and Real-World Examples for Graph, Packet, Sequence, State Diagrams}
    \label{fig:image_examples2}
\end{figure*}

We include the following diagram types in our work: Block, C4, Class Flowchart, Graph, Packet, Sequence, and State diagrams. We provide examples of each diagram type (both synthetic and hand-drawn versions) in Fig. \ref{fig:image_examples1} and Fig. \ref{fig:image_examples2}.


\section{\benchmarkName\ Creation} \label{app:teching-creation}

\begin{figure*}
    \centering
    \includegraphics[width=0.9\linewidth]{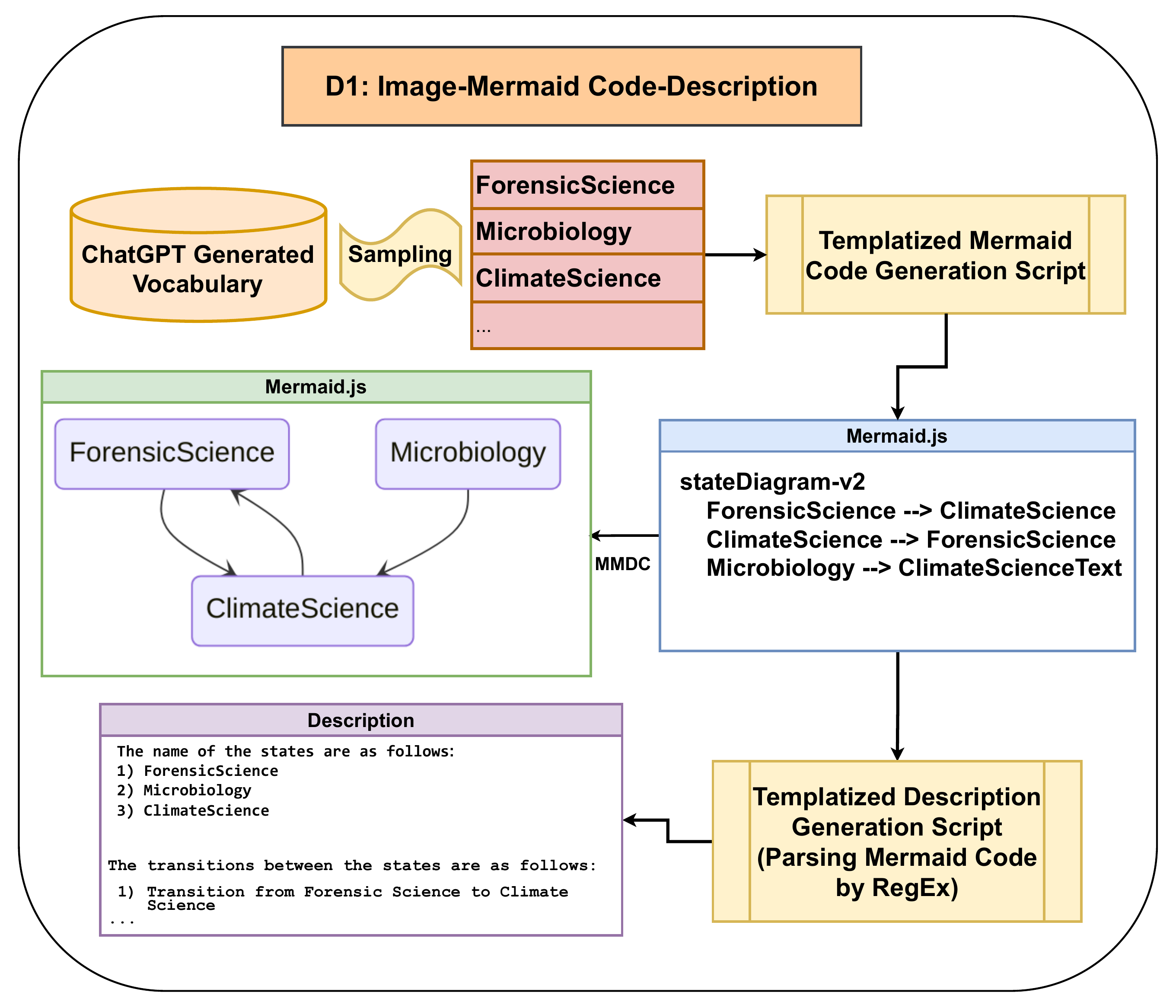}
    \caption{\textbf{D1}: Image-Mermaid Code-Description Corpus generation process.}
    \label{app:fig:D1-Image-Code-pair}
\end{figure*}

\begin{figure*}
    \centering
    \includegraphics[width=0.9\linewidth]{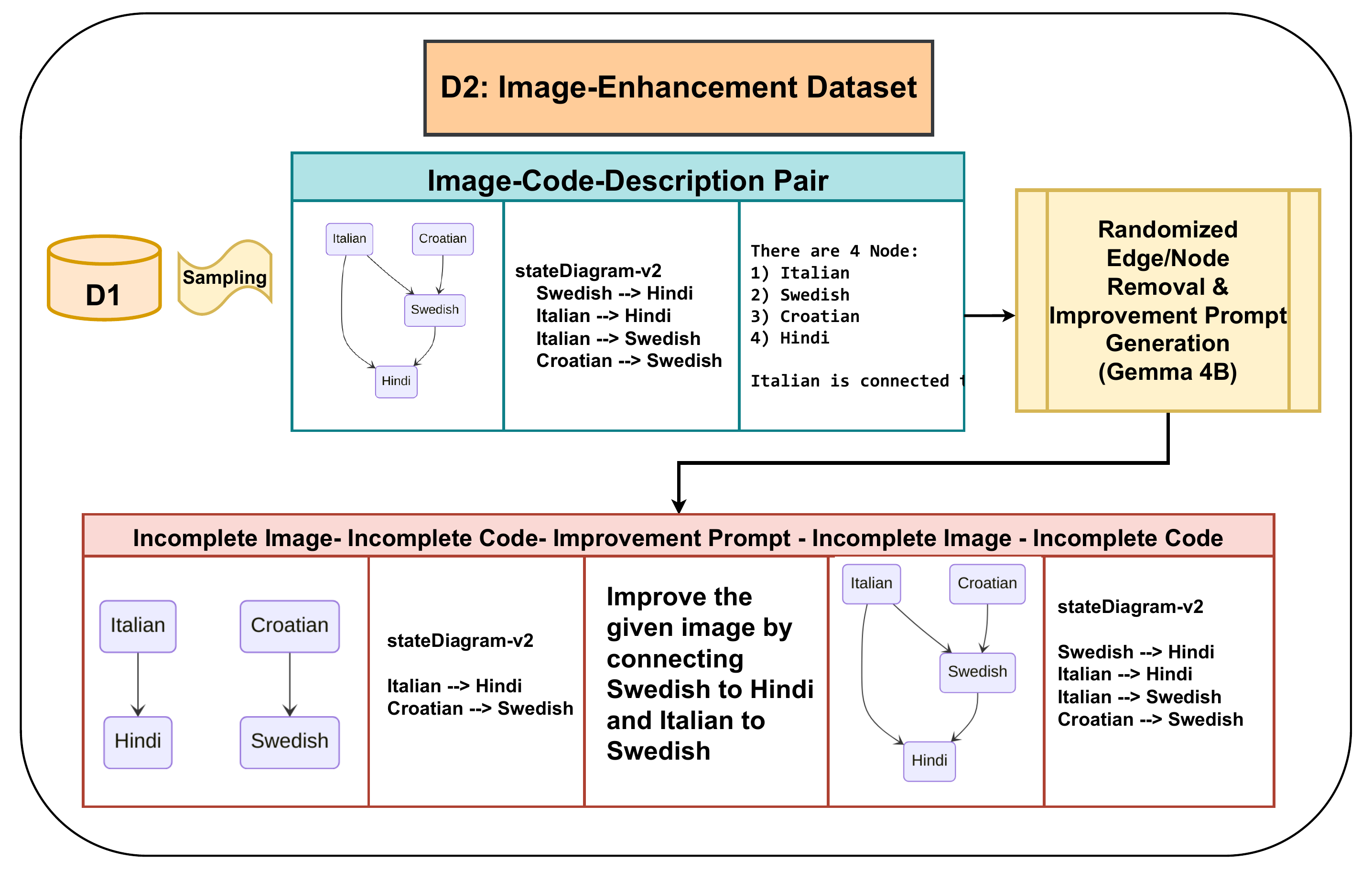}
    \caption{\textbf{D2}: Image Enhancement Corpus generation process} 
    \label{app:fig:D3-Image-enhancement-pair}
\end{figure*}

Fig. \ref{app:fig:D1-Image-Code-pair} and Fig. \ref{app:fig:D3-Image-enhancement-pair} illustrate the dataset creation pipelines for \textbf{D1} (Image-Mermaid Code-Description Corpus) and \textbf{D2} (Image-Enhancement Corpus), providing a visual overview of the step-by-step processes involved in generating each corpus.

\section{\benchmarkName\ Statistics} \label{app:sec-dataset-stats}
Table \ref{tab:diagram_stats} provides the detailed statistics about the various diagram types in the dataset across different difficulty levels (Easy, Medium, and Hard). For each diagram type, the table reports the minimum, mean ± standard deviation, and maximum values for several structural properties, including the number of edges, blocks/classes, attributes, headers, and code length. These statistics highlight the variation in structural complexity across both diagram categories and difficulty levels. For instance, diagrams in the Medium and Hard categories generally exhibit a higher number of edges and blocks, along with longer code lengths, indicating more intricate layouts and richer content. Conversely, Easy diagrams tend to have simpler structures with fewer components, making them less challenging for both model training and evaluation. This detailed breakdown provides a comprehensive view of the dataset’s distribution and structural diversity, which is crucial for understanding model performance across different levels of complexity.

\begin{table*}[ht]
    \centering
    {\fontsize{7.3}{9}\selectfont
    \renewcommand{\arraystretch}{1.2}
    % \begin{adjustbox}{width=\textwidth}
\begin{tabular}{l l l l l l l}
\toprule
\textbf{Diag Type} & \textbf{Level} & \textbf{Edges} & \textbf{Blocks/Classes} & \textbf{Attributes} & \textbf{Headers} & \textbf{Code Length} \\
\midrule
\multirow{3}{*}{\textbf{Block}} & Easy   & (1, 1.00 ± 0.00, 1)        & (2, 2.01 ± 0.11, 3) & - & - & (80, 142.6 ± 28.3, 237) \\
      & Medium & (1, 4.94 ± 0.92, 6) & (4, 4.51 ± 0.78, 7) & - & - & (247, 439.2 ± 112.6, 772) \\
      & Hard   & (1, 6.16 ± 1.73, 9) & (4, 6.05 ± 1.44, 9) & - & - & (303, 568.6 ± 97.8, 971) \\
\midrule
\multirow{3}{*}{\textbf{C4}}    & Easy   & (1, 1.00 ± 0.00, 1)        & (1, 1.50 ± 0.50, 2) & - & - & (221, 314.9 ± 36.4, 415) \\
      & Medium & (4, 4.77 ± 0.66, 6) & (3, 4.52 ± 0.88, 6) & - & - & (833, 1077.2 ± 91.6, 1352) \\
      & Hard   & (4, 4.77 ± 0.67, 6) & (3, 4.52 ± 0.869, 6) & - & - & (859, 1079.6 ± 91.0, 1355) \\
\midrule
\multirow{3}{*}{\textbf{Class}} & Easy   & (1, 1.00 ± 0.00, 1)        & (2, 2.00 ± 0.00, 2) & (4, 4.67 ± 1.56, 10) & - & (198, 254.2 ± 46.8, 459) \\
      & Medium & (3, 3.25 ± 0.66, 5) & (4, 4.25 ± 0.66, 6) & (9, 17.09 ± 3.59, 33) & - & (460, 726.8 ± 133.7, 1287) \\
      & Hard   & (5, 5.00 ± 0.00, 5)        & (6, 6.00 ± 0.00, 6) & (14, 23.92 ± 2.80, 34) & - & (730, 1016.3 ± 89.2, 1291) \\
\midrule
\multirow{3}{*}{\textbf{Flowchart}} & Easy   & (3, 3.50 ± 0.50, 4)  & (4, 4.50 ± 0.50, 5)   & - & - & (173, 222.0 ± 29.7, 277) \\
          & Medium & (4, 6.18 ± 1.40, 9) & (5, 6.00 ± 0.82, 7) & - & - & (196, 287.8 ± 50.7, 394) \\
          & Hard   & (4, 5.67 ± 1.41, 9) & (5, 5.93 ± 0.69, 7) & - & - & (198, 279.7 ± 45.8, 393) \\
\midrule
\multirow{3}{*}{\textbf{Graph}} & Easy   & (1, 1.68 ± 0.47, 2) & (3, 3.34 ± 0.57, 5) & - & - & (41, 109.0 ± 36.9, 234) \\
      & Medium & (3, 3.76 ± 0.83, 5) & (2, 4.06 ± 0.93, 8) & - & - & (62, 141.8 ± 56.9, 344) \\
      & Hard   & (3, 5.64 ± 1.46, 7) & (2, 4.72 ± 1.36, 8) & - & - & (81, 131.2 ± 20.5, 213) \\
\midrule
\multirow{3}{*}{\textbf{Packet}} & Easy   & - & - & - & (2, 2.97 ± 0.17, 3) & (95, 120.8 ± 8.9, 142) \\
       & Medium & - & - & - & (3, 4.99 ± 1.71, 8) & (114, 177.6 ± 49.6, 322) \\
       & Hard   & - & - & - & (6, 8.23 ± 2.14, 17) & (174, 258.7 ± 43.9, 419) \\
\midrule
\multirow{3}{*}{\textbf{Sequence}} & Easy   & (1, 1.44 ± 0.98, 4) & (2, 2.35 ± 0.76, 4) & - & - & (47, 85.7 ± 67.4, 268) \\
         & Medium & (3, 3.50 ± 0.50, 4) & (2, 3.91 ± 0.32, 4) & - & - & (137, 232.1 ± 34.9, 366) \\
         & Hard   & (5, 6.60 ± 1.11, 8) & (6, 7.61 ± 1.10, 9) & - & - & (303, 422.0 ± 67.6, 538) \\
\midrule
\multirow{3}{*}{\textbf{State}} & Easy   & (1, 1.88 ± 0.86, 4)  & (2, 3.40 ± 1.28, 8) & - & - & (32, 71.8 ± 30.0, 217) \\
      & Medium & (5, 7.81 ± 1.78, 12) & (5, 10.61 ± 2.43, 19) & - & - & (118, 248.9 ± 73.5, 518) \\
      & Hard   & (5, 6.52 ± 0.94, 8)  & (3, 8.33 ± 1.56, 13) & - & - & (441, 703.2 ± 119.8, 1093) \\
\bottomrule
\end{tabular}
\caption{Detailed statistics of different diagram types across difficulty levels. Each cell shows (min, mean ± std, max).}
\label{tab:diagram_stats}
}
\end{table*}

%%%%%%%%%%%%%%%%%%%%%%%%%%%%%

%%%%%%%%%%%%%%%%%%%%%%%%%%%%%

\section{Tasks Descriptions and Examples} \label{app:sec-tasks-examples}
\begin{figure*}[t]
            \centering
            \includegraphics[scale=0.14]{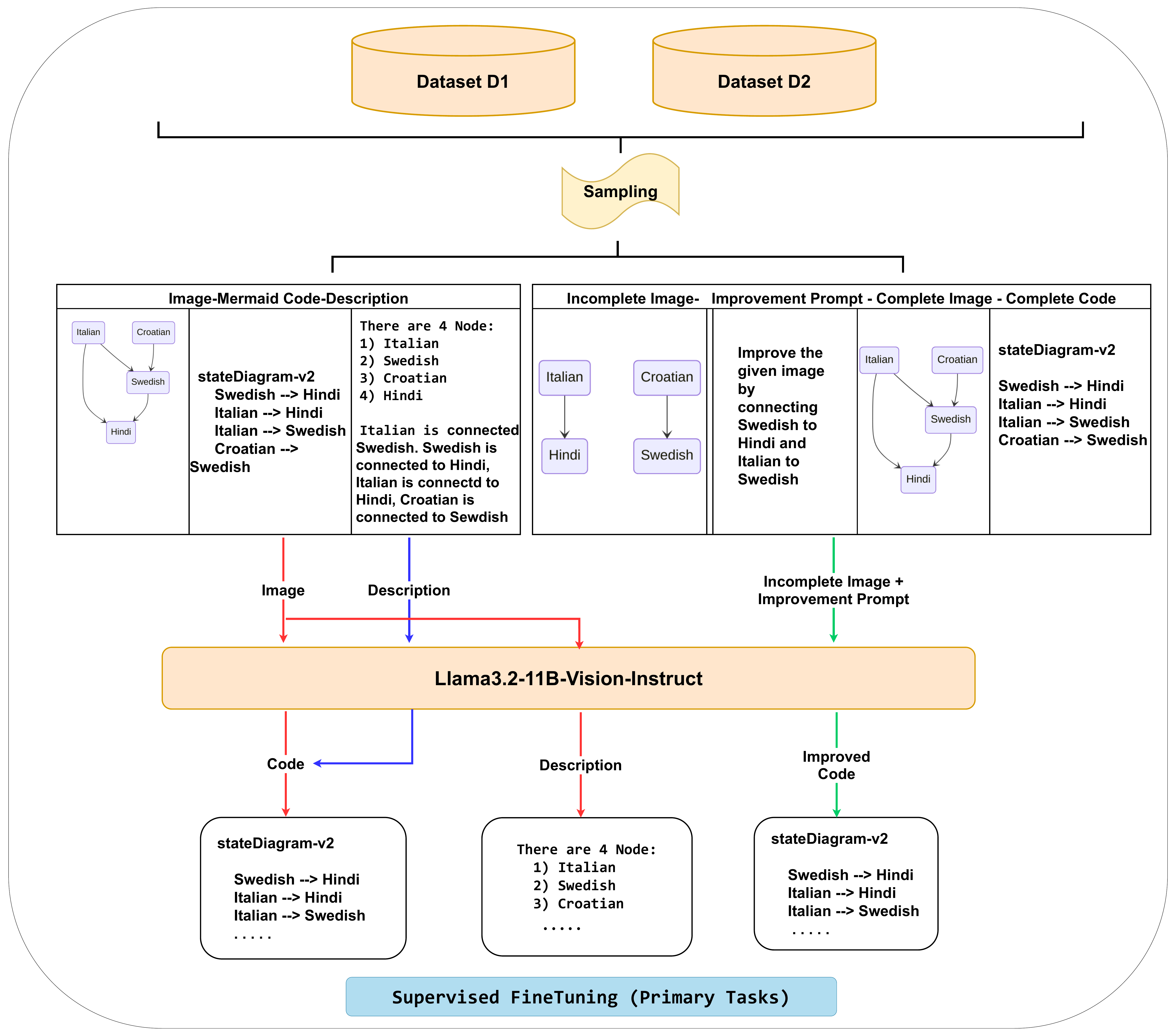                    
  }
            \caption{Diagram illustrating formulation of different Primary Tasks utilizing \benchmarkName\ for finetuning}
            \label{fig:tasks-pipeline}
            %\vspace{-5mm}
\end{figure*}

\begin{figure}[H]
    \centering
    \includegraphics[width=1.0\linewidth]{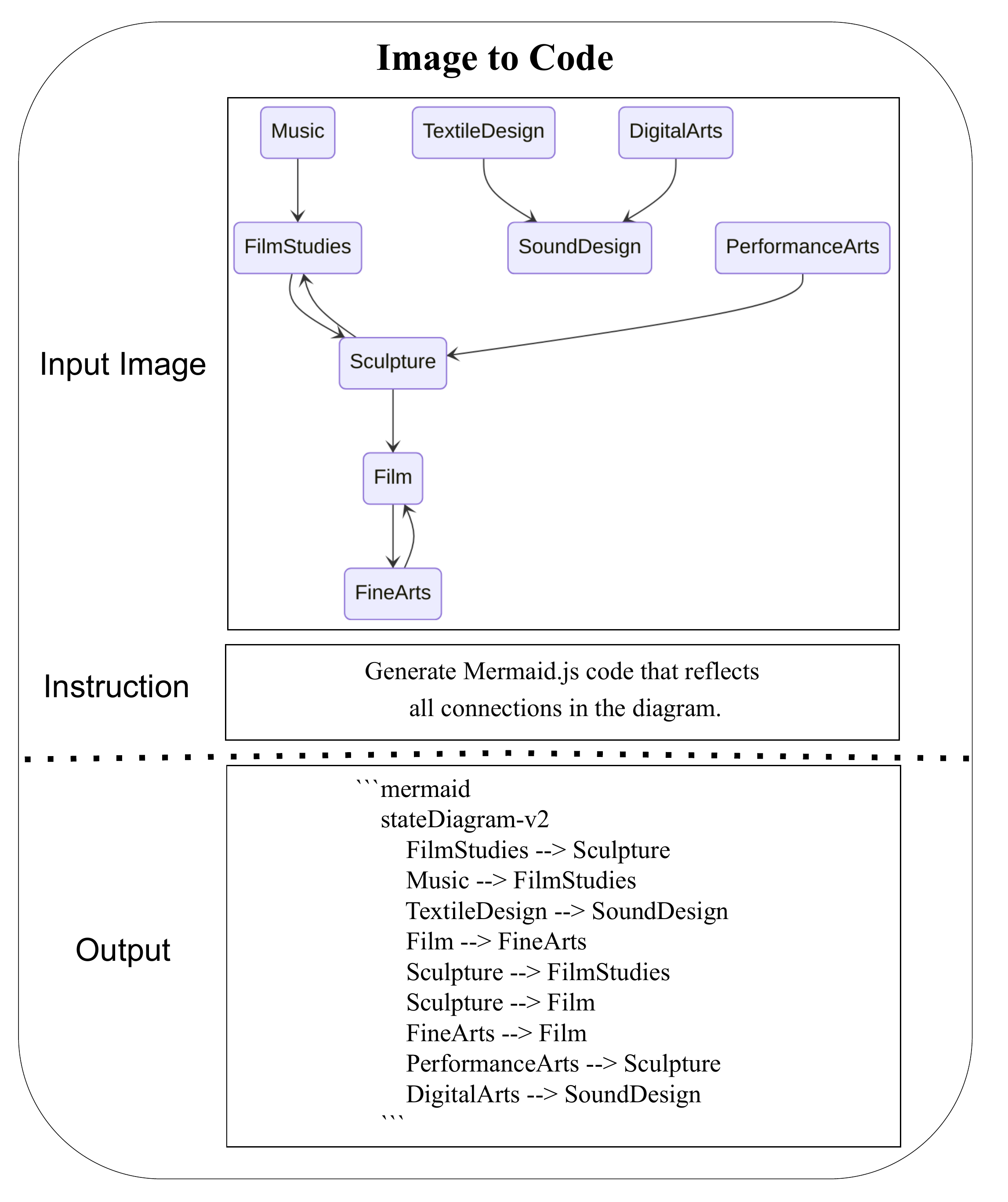}
    \caption{Image2Code Task Example}
    \label{fig:task_image2code}
\end{figure}

\begin{figure}[H]
    \centering
    \includegraphics[width=1.0\linewidth]{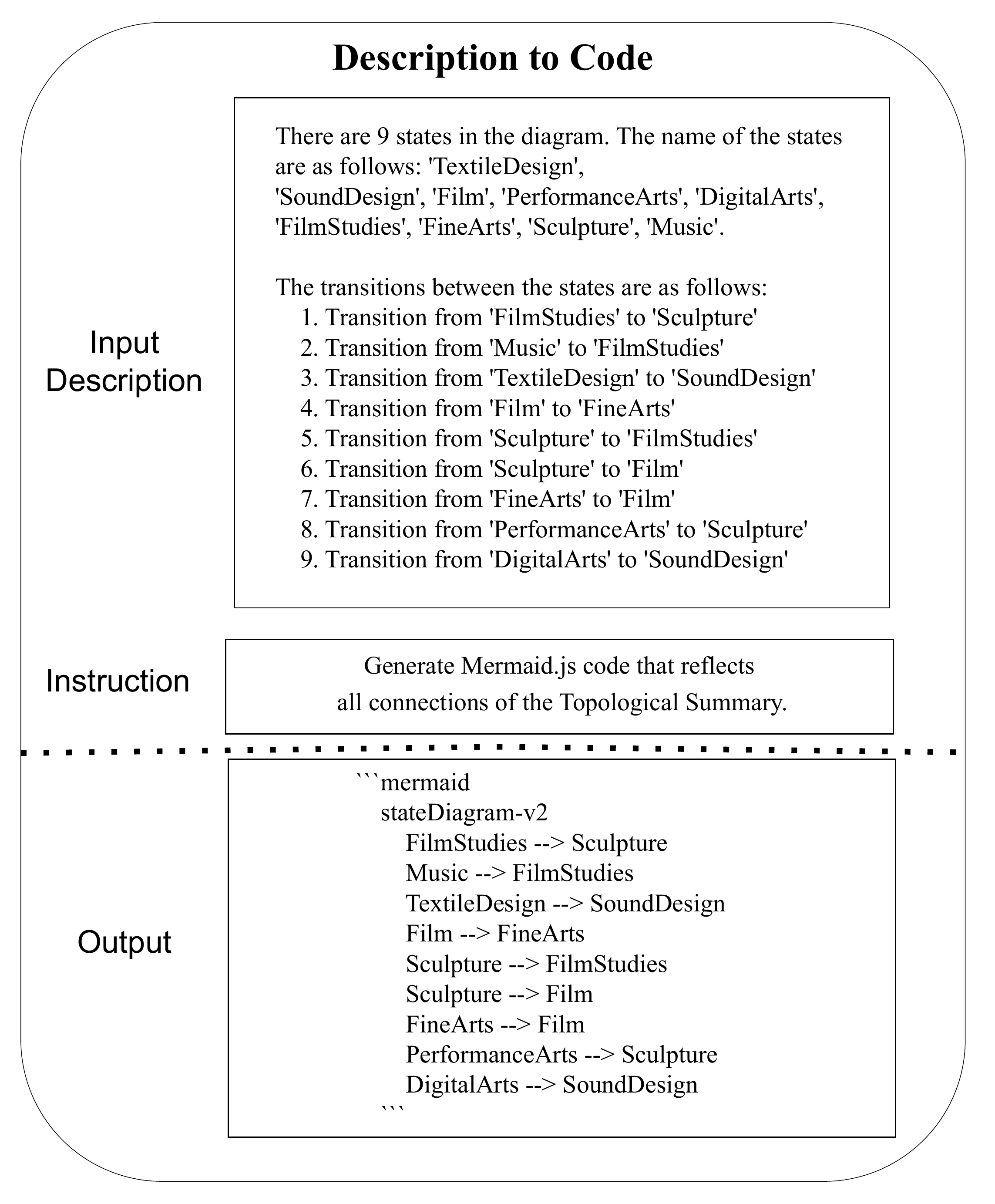}
    \caption{Description2Code Task Example}
    \label{fig:task_description2code}
\end{figure}

\begin{figure}[H]
    \centering
    \includegraphics[width=0.9\linewidth]{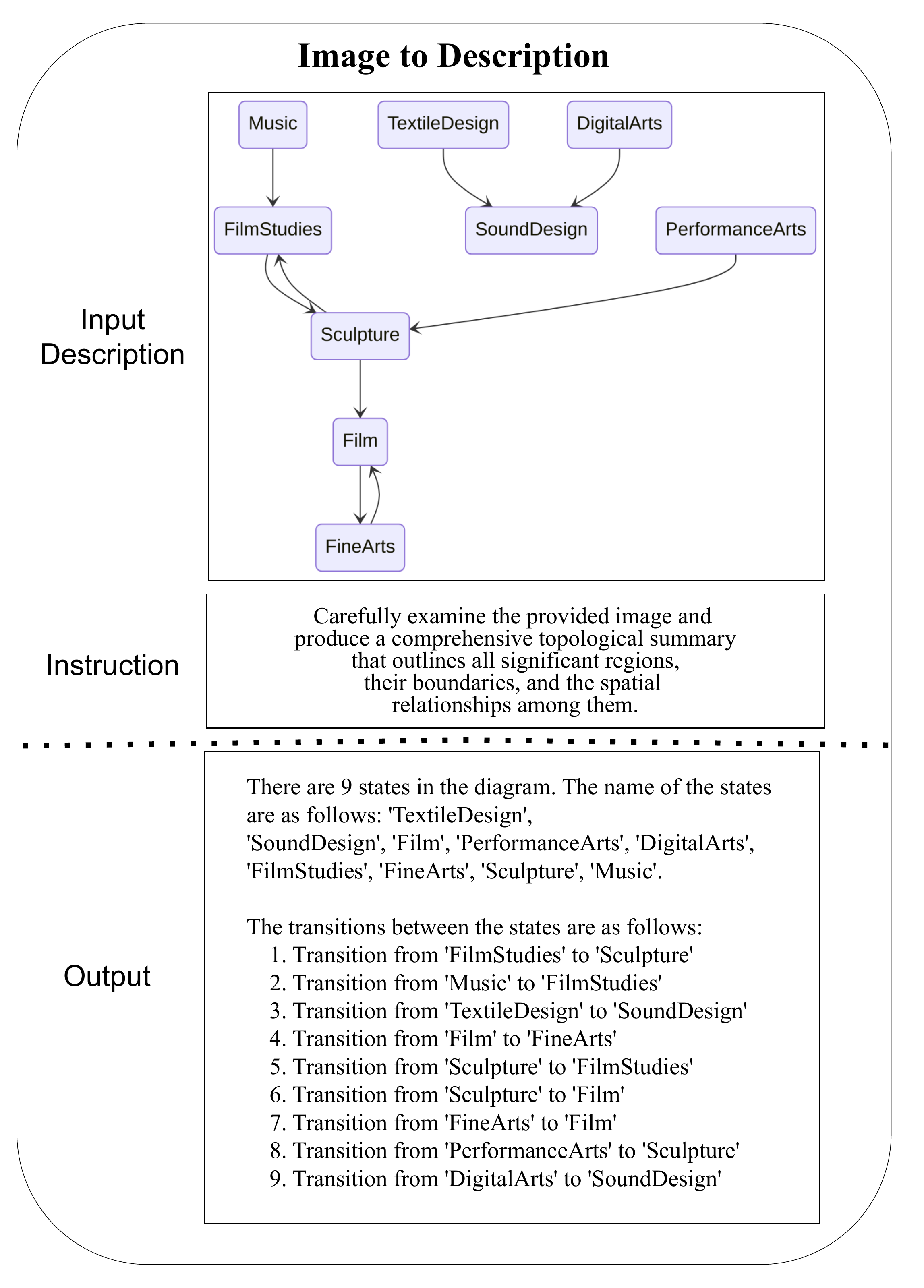}
    \caption{Image2Description Task Example}
    \label{fig:task_image2description}
\end{figure}

\begin{figure}[H]
    \centering
    \includegraphics[width=0.9\linewidth]{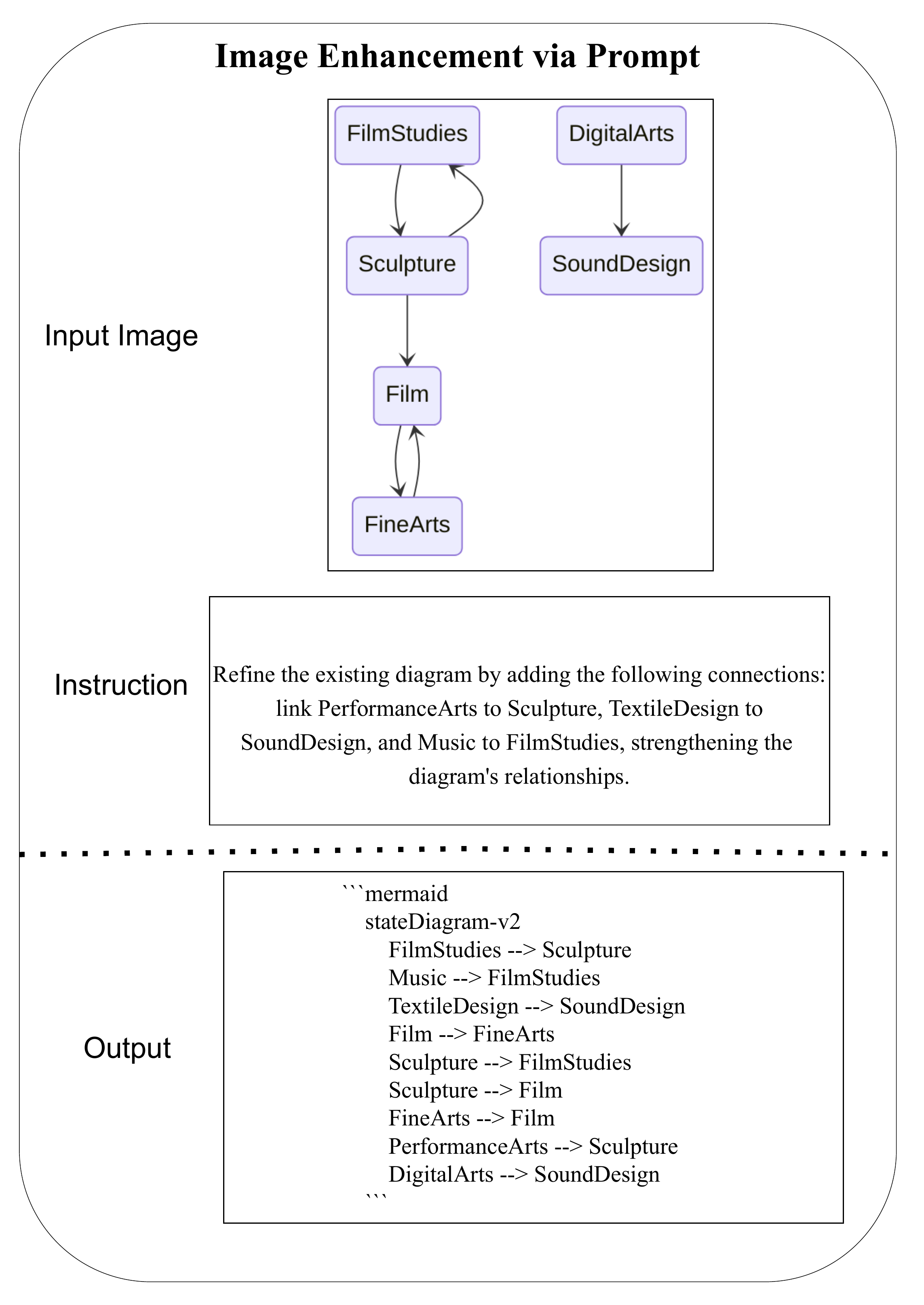}
    \caption{Example of Image Enhancement via Prompt}
    \label{fig:task_imageenhanceviaprompt}
\end{figure}

The primary task is described in detail in the main paper. Fig. \ref{fig:tasks-pipeline} shows the formulation of different Primary Tasks utilizing \benchmarkName. Fig. \ref{fig:task_image2code}, \ref{fig:task_description2code}, \ref{fig:task_image2description} and \ref{fig:task_imageenhanceviaprompt} provides examples of each of the primary tasks, Image2Code, Description2Code, Image2Description, and Image Enhancement via Prompt. The self-supervision tasks, which are used exclusively during training to improve fine-grained alignment and structural understanding, are outlined below. 

\begin{figure}[H]
    \centering
    \includegraphics[width=1.0\linewidth]{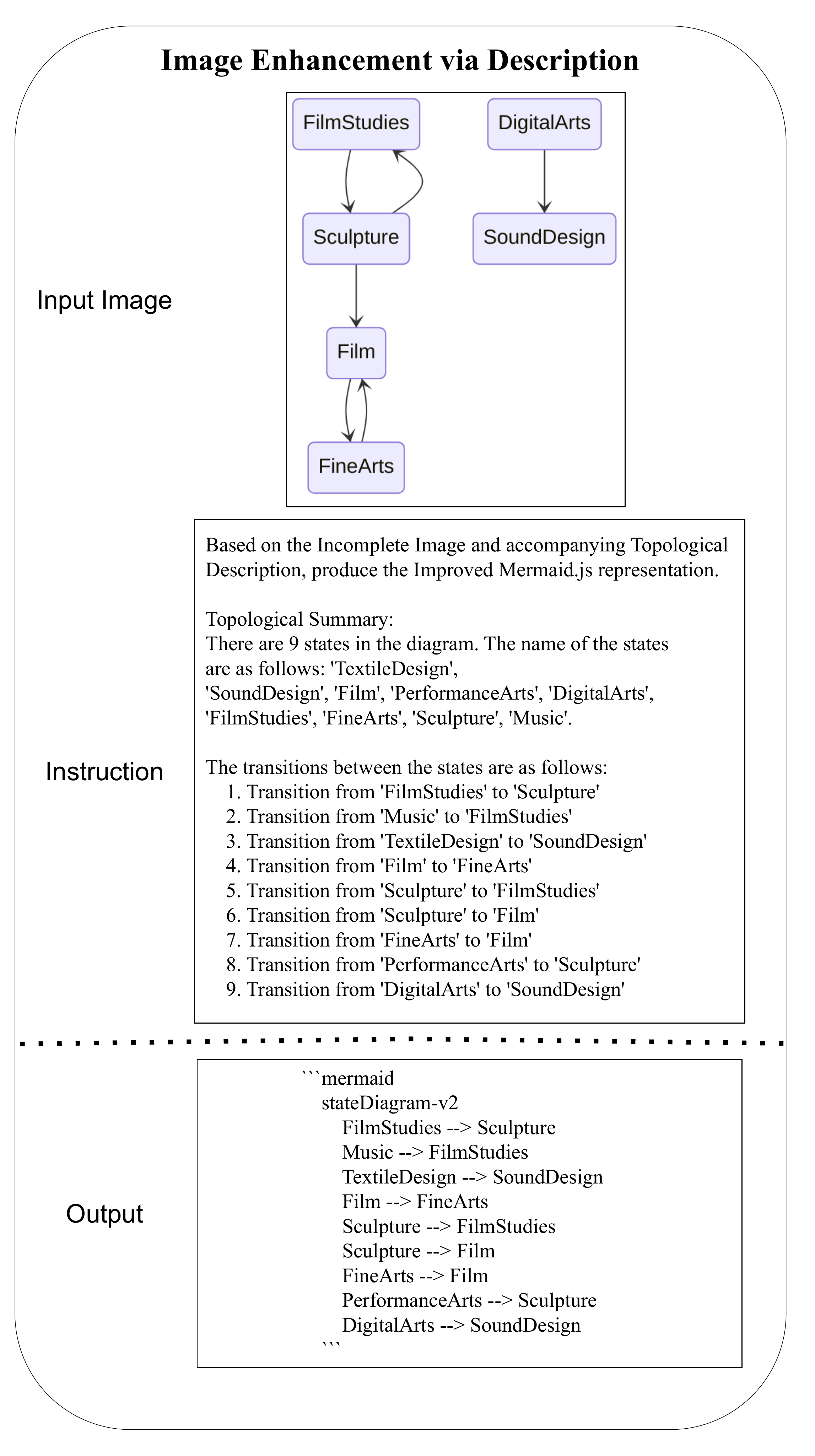}
    \caption{Example of Image Enhancement via Description}
    \label{fig:task_imageenhanceviadescription}
\end{figure}

\noindent\textbf{{Image Enhancement via Description}:} In this task, the model is given an image along with a textual description of the target image, and must produce code that reflects the enhanced description, effectively adding missing components to align with the full target diagram. By leveraging both visual and textual cues, the model learns fine-grained alignment between the image, description, and code, enabling it to enhance incomplete diagrams and generate more complete and accurate code representations. This training signal also enables the model to recover missing parts in diagram components and relationships by grounding textual edits in the visual structure of the input image. An example of the task is shown in Fig. \ref{fig:task_imageenhanceviadescription}.

\begin{figure}[H]
    \centering
    \includegraphics[width=1.0\linewidth]{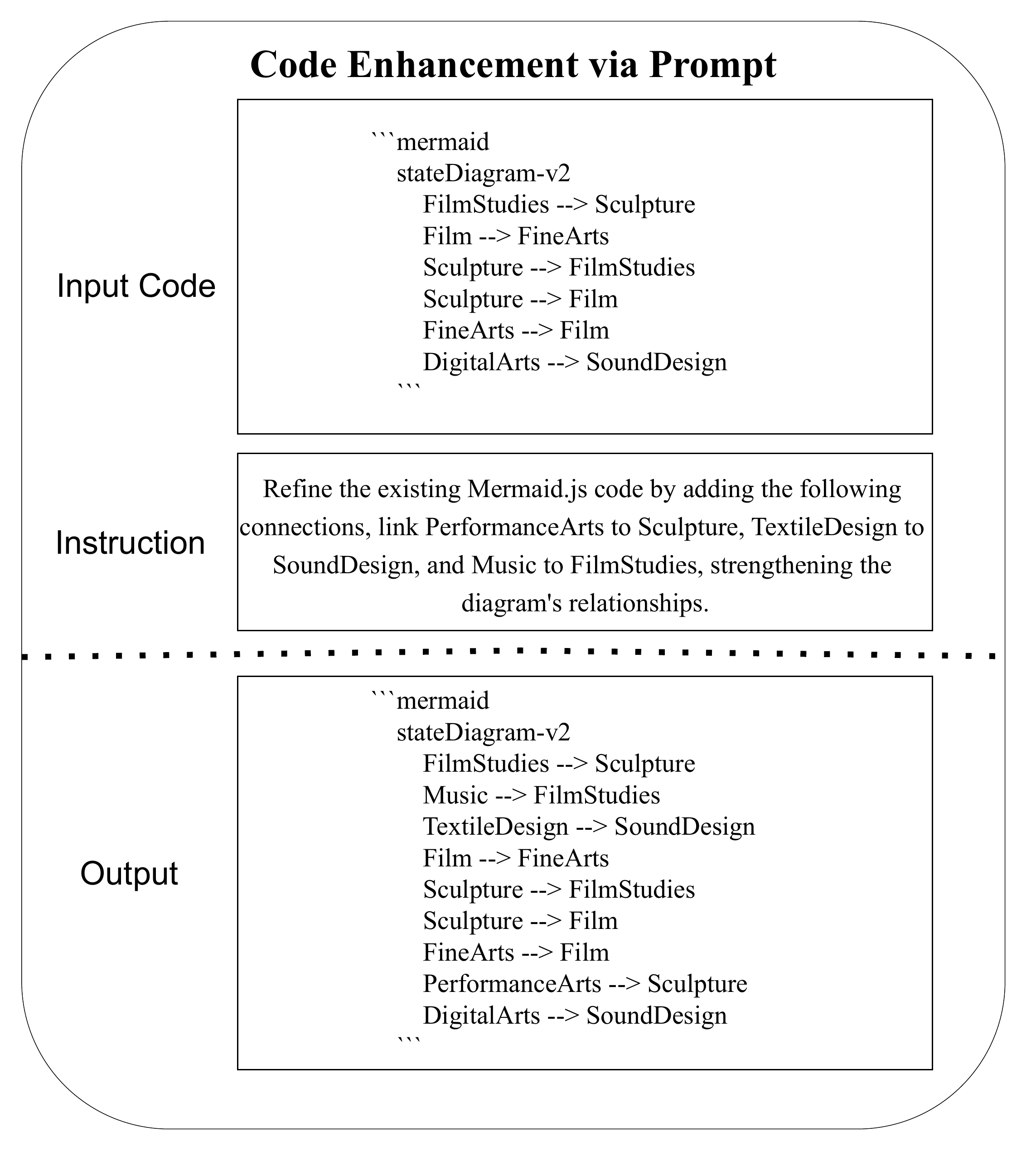}
    \caption{Example of Code Enhancement via Prompt}
    \label{fig:task_codeenhanceviaprompt}
\end{figure}

\noindent\textbf{{Code Enhancement via Prompt}:} The model is given existing Mermaid code along with an enhancement prompt and must update the code accordingly. This encourages the model to understand structural modifications in the provided instructions via prompt and apply contextual edits, which is crucial for handling incremental edits and refinements during diagram generation and editing. Fig. \ref{fig:task_codeenhanceviaprompt} shows a representative example of the task.

\begin{figure}[H]
    \centering
    \includegraphics[width=1.0\linewidth]{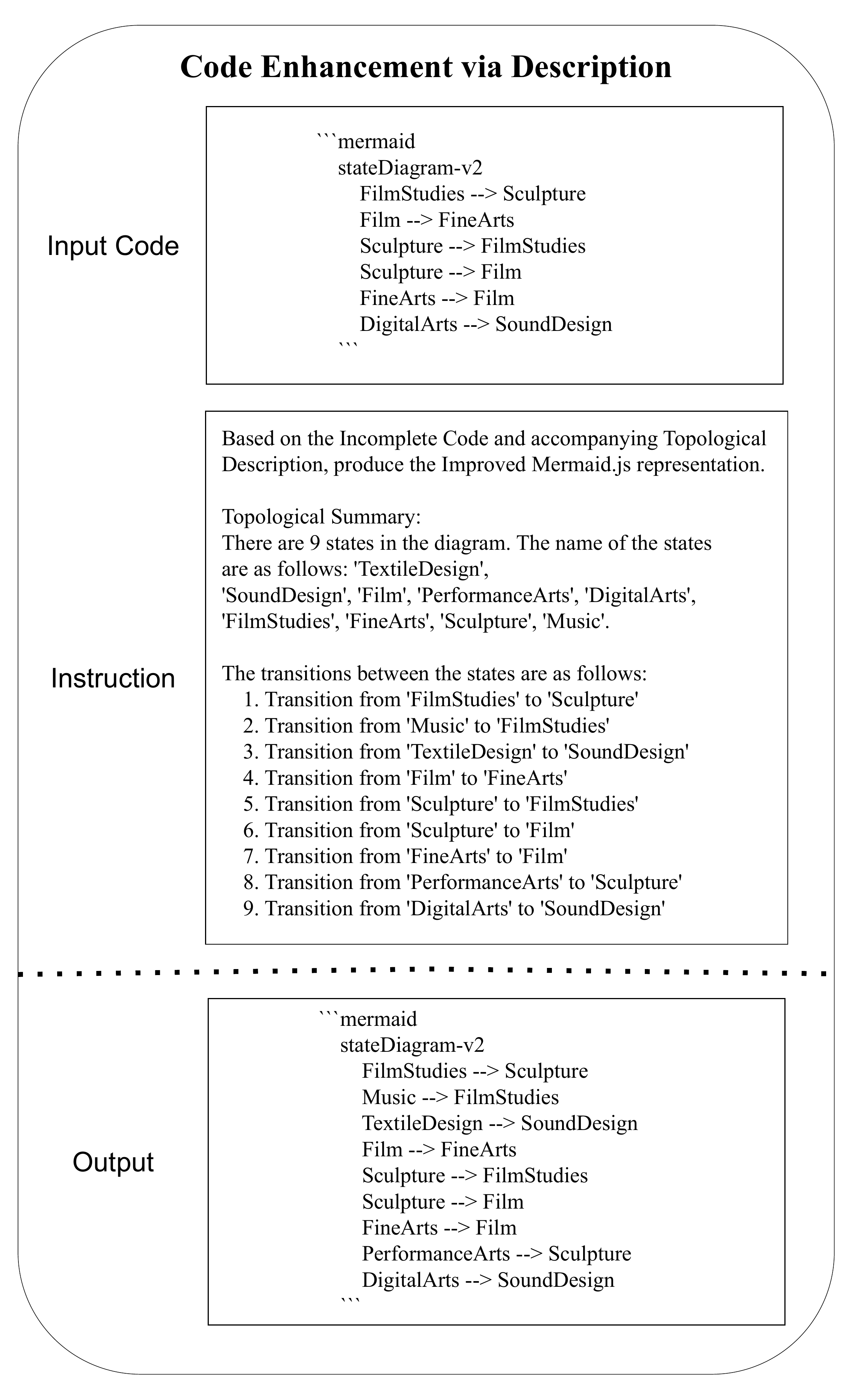}
    \caption{Example of Code Enhancement via Description}
    \label{fig:task_codeenhanceviadescription}
\end{figure}

\noindent\textbf{{Code Enhancement via Description}:} The model receives a Mermaid code snippet along with a natural language description of the target image and is tasked with enhancing the code to accurately reflect the changes present in the description. This objective requires the model to interpret textual description of the image and apply the corresponding structural modifications to the code. As a result, the task strengthens code completion and generation capabilities, ensuring the model can effectively bridge gaps between natural language descriptions and structured code representations. An illustration of the task can be seen in Fig. \ref{fig:task_codeenhanceviadescription}.

\begin{figure}[H]
    \centering
    \includegraphics[width=1.0\linewidth]{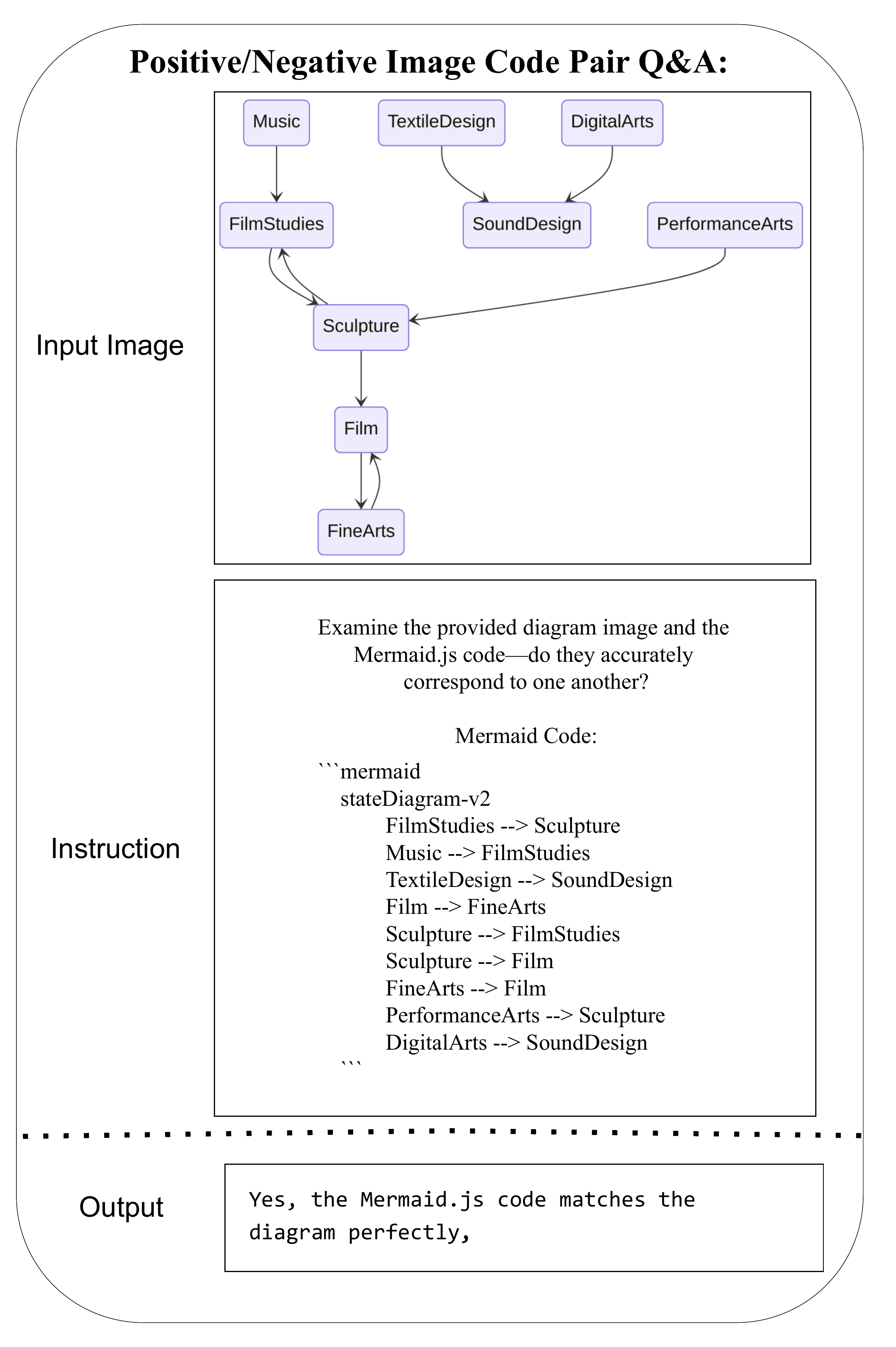}
    \caption{Example of Positive/Negative Image Code pair Q/A}
    \label{fig:task_positive_image_code_qa}
\end{figure}

\noindent\textbf{{Positive/Negative Image–Code Pair Q\&A}:} The model predicts whether a given image–code pair constitutes a valid match or a mismatch. This binary classification objective requires the model to jointly reason over visual and structural code representations, enabling it to distinguish semantically consistent pairs from incorrect or misaligned ones. By explicitly learning to identify mismatches, the model develops a stronger notion of cross-modal alignment, which improves robustness to noisy or ambiguous inputs. Fig. \ref{fig:task_positive_image_code_qa} illustrates an example of the task.

\begin{figure}[H]
    \centering
    \includegraphics[width=1.0\linewidth]{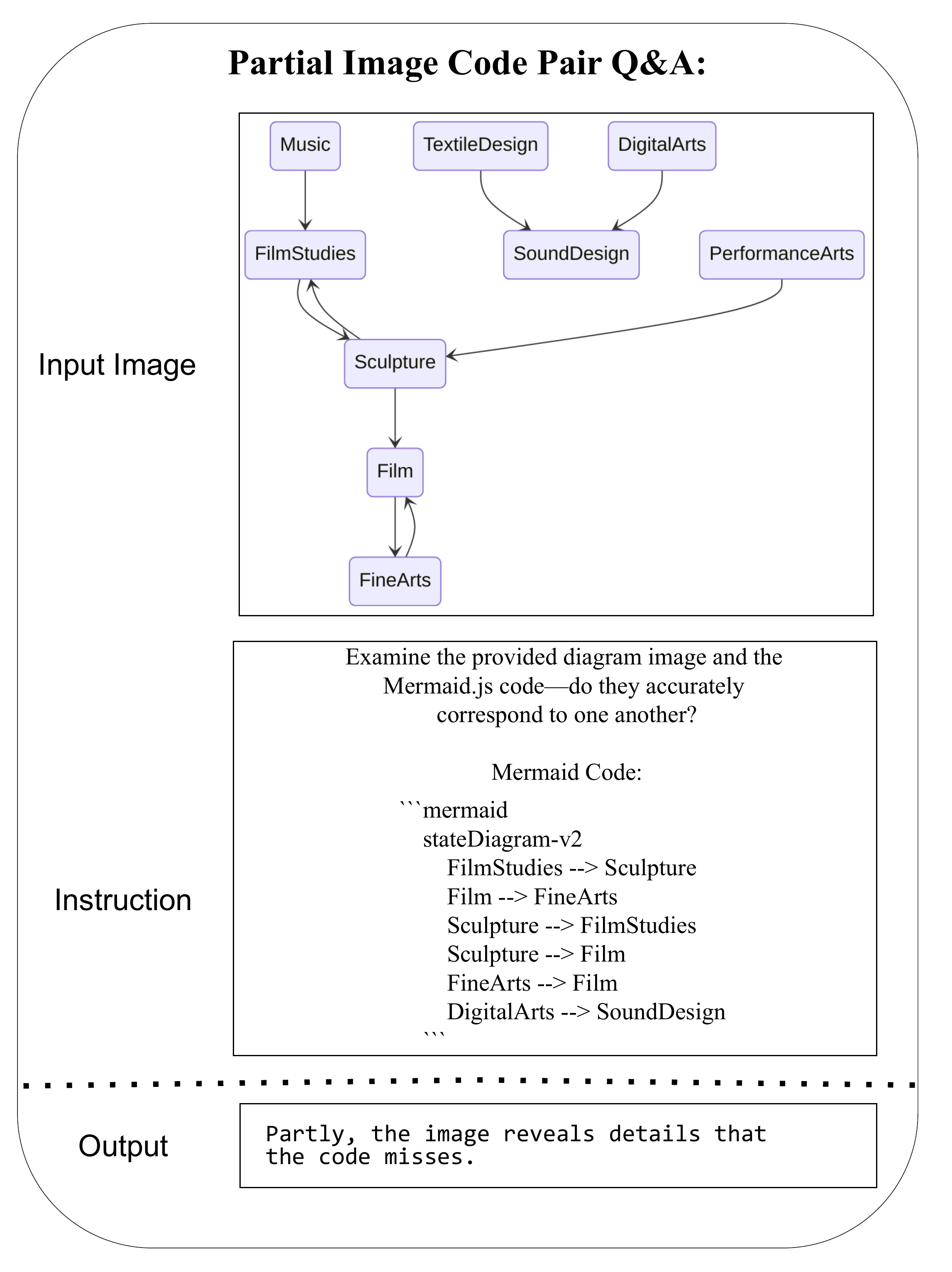}
    \caption{Example of Partial Image Code pair Q/A}
    \label{fig:task_partial_image_code_qa}
\end{figure}

\noindent\textbf{{Partial Match Image–Code Pair Q\&A}:} The model determines whether the generated code partially matches the given image by capturing fine-grained sub-part relationships between visual elements and their corresponding code components. Instead of enforcing a complete one-to-one correspondence, this formulation encourages alignment between relevant subsets of the image and portions of the code, even when the match is incomplete. Through this process, the model learns to reason about partial correspondences between visual components and structured code representations, which is particularly beneficial for complex, noisy, or incomplete diagrams. Fig. \ref{fig:task_partial_image_code_qa} depicts how the task is performed through an example.

%%%%%%%%%%%%%%%%%%%%%%%%%%%%%%%

\section{Baseline Models} \label{app:sec:baseline-models}
We had used several baseline Multimodal LLMs which are of the similar parameter size of our base model Llama3.2-11B-Vision-Instruct \cite{Llama3.2-Vision(11B)}. We experimented with MiniCPM-V-2-6(8B) \cite{yao2024minicpmvgpt4vlevelmllm}, Qwen2.5-VL-7B-Instruct \cite{bai2023qwenvlversatilevisionlanguagemodel}, Gemma3-12B-Instruction-Tuned \cite{gemma3} and GPT-4o-mini \cite{openai_gpt4o_mini}

%%%%%%%%%%%%%%%%%%%%%%%%%%%%%%%

\section{Augmentations} \label{app:sec-augmentations}

\begin{figure*}[t]  % use figure* instead of figure
    \centering
    \includegraphics[width=1.0\textwidth]{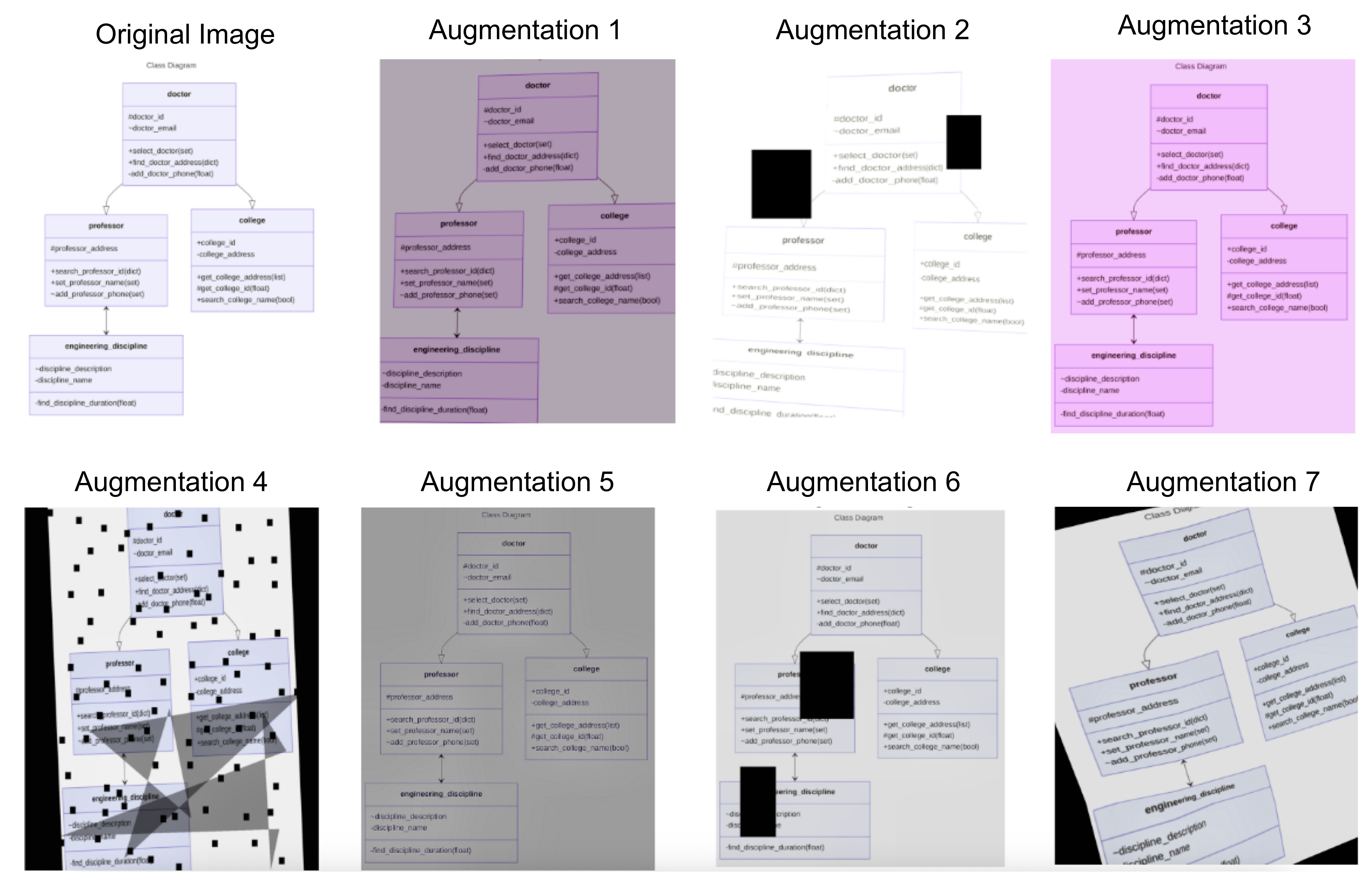}
    \caption{Examples of Augmentations used for keeping diversity and preventing overfitting.}
    \label{fig:augmentation}
\end{figure*}

Data augmentation is a strategy to artificially increase the diversity of the  training dataset by applying transformations. The augmentation pipeline we employ plays a crucial role in improving the model’s ability to generalize across real-world imaging conditions. By introducing a diverse set of transformations, including blur, noise, lighting changes, geometric distortions, and weather effects, the model is exposed to the kinds of variability commonly encountered in natural settings, such as those captured by camera sensors. For instance, motion blur and Gaussian noise simulate camera shake or low-light grain, while color jitter and brightness/contrast shifts mimic different lighting environments. Perspective transforms and random rotations help the model become invariant to viewpoint changes, making it robust to varied camera angles or object orientations. Similarly, occlusion-based augmentations like coarse or grid dropout encourage the model to learn from partial visual cues, improving resilience when parts of the image are obscured. Collectively, these augmentations ensure that the model doesn't overfit to clean, synthetic data but instead learns features that are stable and discriminative under real-world noise and viewpoint variations. For the augmentation procedure, we follow the steps shown in Algorithm~\ref{alg:augmentation}. We use the Albumentations library \url{https://albumentations.ai/}. Fig. \ref{fig:augmentation} illustrates some examples of augmentations used in the training process.

\begin{algorithm*}[ht]
\caption{Image Augmentation Pipeline}
\label{alg:augmentation}
\begin{algorithmic}[1]
\State \textbf{Input:} Image $I$
\State \textbf{Output:} Augmented image $I'$
\Procedure{Augment}{$I$}
    \State Select one (with probability $p = 0.7$):
    \State \quad Motion Blur ($p = 0.5$)
    \State \quad Median Blur (limit = 3, $p = 0.5$)
    \State \quad Gaussian Blur (limit = [3, 5], $p = 0.5$)
    \State Select one (with probability $p = 0.9$):
    \State \quad RGB Shift (limit = 80, $p = 0.8$)
    \State \quad Hue/Saturation/Value Shift (limits: 15/25/20, $p = 0.8$)
    \State Apply Random Brightness/Contrast (brightness = 0.4, contrast = 0.2, $p = 0.7$)
    \State Select one (with probability $p = 0.3$):
    \State \quad Rotate (limit = 30°, $p = 0.5$)
    \State \quad Shift/Scale/Rotate (shift = 0.1, scale = 0.2, rotate = 45°, $p = 0.3$)
    \State Apply Grid Distortion ($p = 0.5$) with overall probability $p = 0.2$
    \State Apply Perspective Transform (scale = [0.01, 0.05], $p = 0.4$)
    \State Select one:
    \State \quad Coarse Dropout (holes = 4, size = 16×16, $p = 0.3$)
    \State \quad Grid Dropout (ratio = 0.2, $p = 0.1$)
    \State Select one (with probability $p = 0.4$):
    \State \quad Random Rain ($p = 0.3$)
    \State \quad Random Fog ($p = 0.3$)
    \State \quad Random Shadow ($p = 0.8$)
    \State Optionally apply CLAHE ($p = 0.3$)
    \State Optionally apply Sharpen ($p = 0.3$)
    \State \textbf{return} Augmented image $I'$
\EndProcedure
\end{algorithmic}
\end{algorithm*}

\section{Training and Hyperparameters Details} \label{app:sec:hyperparam}

We performed supervised finetuning of Llama-3.2-11B-Vision-Instruct \cite{Llama3.2-Vision(11B)} using parameter-efficient finetuning method LoRA \cite{lora} using hyperparameters outlined in Table \ref{tab:hyperparameters}. Overall finetuning took 88 hours on 2 Nvidia A6000 GPUs.
\begin{table}[H]
    \centering
    % \small
    \renewcommand{\arraystretch}{1.2}
    \begin{tabular}{l l}
        \toprule
        \textbf{Hyperparameters} & \textbf{Value} \\
        \midrule
        per\_device\_train\_batch\_size & 1 \\
        gradient\_accumulation\_steps & 1 \\
        learning\_rate & 2e-5 \\
        weight\_decay & 0.05 \\
        num\_train\_epochs & 2 \\
        lr\_scheduler\_type & cosine \\
        warmup\_ratio & 0.2 \\
        bf16 & True \\
        lora rank & 32 \\
        lora\_alpha & 16 \\
        target\_modules & QKV\\
        lora\_dropout & 0.2 \\
        use\_rslora & True \\
        \bottomrule
    \end{tabular}
    \caption{Training Hyperparameters.}
    \label{tab:hyperparameters}
\end{table}

\section{Evaluation Metrics} \label{app:sec-metrics}

We employ a combination of standard automatic language-based metrics and human evaluation metrics with custom structural heuristics to evaluate the performance of models across synthetic and real-world datasets as presented in Table \ref{tab:metrics}.

\begin{table}[H]
    \centering
    \small
    \renewcommand{\arraystretch}{1.2}
    \resizebox{0.5\textwidth}{!}{%
    \begin{tabular}{l l}
        \toprule
        \textbf{Task} & \textbf{Metrics} \\
        \midrule
        \multirow{6}{*}{\shortstack[l]{\textbf{Automatic Evaluation}\\Image2Code (Synthetic) \\Description2Code (Synthetic) \\Image2Description (Synthetic)\\Image2code (Realworld)}}
         & BLEU \\
         & SACREBLEU \\
         & METEOR \\
         & chrF \\
         & BLEURT  \\
         & ROUGE-L \\ 
         \midrule
         \multirow{6}{*}{\shortstack[l]{\textbf{Human Evaluation}\\Image2code (Realworld)}}
         &  Compilation Error \\
         & Correct Blocks \\
         & Correct Edges \\
         & Correct Labeled Edges \\
         & Correct Attributes \& Methods \\
         & Correct Bits  \\
         \bottomrule
    \end{tabular}%
    }
    \caption{Evaluation metrics}
    \label{tab:metrics}
\end{table}

\subsection{Automatic Evaluation Metrics} \label{app:sec-auto-metrics} 

For Image2Code, Description2Code, and Image2Description tasks on synthetic data, where the ground truth Mermaid code is clean and perfectly aligned with the diagram, we use established text similarity metrics i.e., BLEU, SACREBLEU, METEOR, chrF, BLEURT, and ROUGE-L. These metrics quantify lexical and semantic overlap between the generated and ground truth Code/Description, providing a baseline for textual accuracy.

\subsection{Human Evaluation Metrics} \label{app:sec-human-metrics}
For the Image2Code task on Realworld Images Corpus \textbf{D3}, the evaluation focuses on both structural fidelity between the generated code and the original diagram. Correct Blocks/Correct Attributes and Methods/Correct Bits measures how accurately the model identifies and reproduces the diagram’s components, depending on the diagram type. Correct Edges assesses the accuracy of the relationships or connections between these components. Correct Labeled Edges adds an additional layer by verifying whether the labels on the connections are also correctly generated. Finally, Compilation Error identifies whether the generated code is syntactically and semantically correct enough to compile by the Mermaid compiler without errors.

\vspace{-1mm}
\section{Detailed Experimental Results} \label{app:section-I}
\vspace{-1mm}
This section presents the complete set of experimental results in tabular form (Tables \ref{tab:image_2_code_metrics}, \ref{tab:summary_2_code_metrics}, \ref{tab:topological_summary_metrics}, and \ref{tab:auto_real_world}). These include performance metrics across all diagram types and models, complementing the summary findings discussed in the main paper. The tables provide a more granular view for in-depth analysis. Tables \ref{tab:image_2_code_metrics}, \ref{tab:summary_2_code_metrics}, \ref{tab:topological_summary_metrics}, and \ref{tab:auto_real_world} present the detailed automatic evaluation results for the Image2Code, Description2Code, and Image2Description tasks on synthetic images, as well as the Image2Code task on real-world images, respectively. 

\noindent Table \ref{tab:old_ablation} details the preliminary results obtained in a continual training setup where model  trained sequentially on Image2Code, then Description2Code, and finally on a self-supervised image-enhancement task showed that performance on the previous primary tasks remained largely unchanged after training on next primary task. In the table, CTS1, CTS2, and CTS3 correspond to sequential stages of model training: CTS1 represents the model fine-tuned on the first primary task (Image2Code); CTS2 represents the model after subsequent fine-tuning on the Description2Code and Code2Description tasks; and CTS3 represents the model after final fine-tuning on the Image Enhancement task.

\begin{table*}[ht]
    \centering
    {\fontsize{7}{9}\selectfont
    \renewcommand{\arraystretch}{1.2}
    % \begin{adjustbox}{width=\textwidth}
    \begin{tabular}{llcccccc}
        \toprule
        \textbf{Diagram Type} & \textbf{Model} & \textbf{BLEU$\uparrow$} & \textbf{SACREBLEU$\uparrow$} & \textbf{METEOR$\uparrow$} & \textbf{chrF$\uparrow$} & \textbf{BLEURT$\uparrow$} & \textbf{ROUGE-L$\uparrow$} \\
        \midrule
        \multirow{6}{*}{\textbf{Block}}          
        & \textbf{Llama3.2-11B-Vision-Instruct} & 0.0059   & 1.9446   & 0.1823   & 27.9335  & -0.6176  & 0.2278   \\
        & \textbf{Llama-VL-TUG (ours)} & \textbf{0.8676} & \textbf{86.7621}    & \textbf{0.9432} & \textbf{94.6115}  & \textbf{0.3255}  & \textbf{0.8953}    \\ 
        & \textbf{Gemma3-12B-Instruction-Tuned} & 0.0488   & 5.6815   & 0.2353   & 32.8237  & -0.5744  & 0.2474   \\
        & \textbf{Qwen2.5-VL-7B-Instruct} & 0.0148   & 3.0378   & 0.2211   & 24.8280  & -0.6823  & 0.2275   \\
        & \textbf{MiniCPM-V-2-6} & 0.0000   & 0.4034   & 0.0669   & 2.9787   & -1.5357  & 0.0231   \\
        & \textbf{GPT-4o-mini} & 0.0021   & 2.4243   & 0.1960   & 18.9216  & -0.7492  & 0.2077   \\ \midrule
        \multirow{6}{*}{\textbf{C4}}
        & \textbf{Llama3.2-11B-Vision-Instruct} & 0.0525   & 5.5587   & 0.1569   & 37.4066  & -0.4687  & 0.3580   \\
        & \textbf{Llama-VL-TUG (ours)} & \textbf{0.9585} & \textbf{95.8483} &   \textbf{0.9845} & \textbf{98.9167} & \textbf{0.2863}  & \textbf{0.9877}     \\ 
        & \textbf{Gemma3-12B-Instruction-Tuned} & 0.0570   & 6.2295   & 0.2171   & 42.5174  & -0.4602  & 0.4325   \\
        & \textbf{Qwen2.5-VL-7B-Instruct} & 0.0657   & 6.9249   & 0.1978   & 39.0591  & -0.5121  & 0.3557   \\
        & \textbf{MiniCPM-V-2-6} & 0.0000   & 0.0322   & 0.0073   & 1.8335   & -1.3834  & 0.0298   \\
        & \textbf{GPT-4o-mini} & 0.1022   & 10.4744  & 0.2325   & 44.4395  & -0.4793  & 0.4898   \\ \midrule
        \multirow{6}{*}{\textbf{Class}}
        & \textbf{Llama3.2-11B-Vision-Instruct} & 0.6317   & 63.1853  & 0.5364   & 75.3705  & 0.1994   & 0.5912   \\
        & \textbf{Llama-VL-TUG (ours)} & \textbf{0.9465} & \textbf{94.6490}   & \textbf{0.9523} & \textbf{99.0225} & 0.2890  & \textbf{0.8181}   \\ 
        & \textbf{Gemma3-12B-Instruction-Tuned} & 0.6328   & 63.2811  & 0.5355   & 75.4252  & 0.1044   & 0.6053   \\
        & \textbf{Qwen2.5-VL-7B-Instruct} & 0.5923   & 59.2725  & 0.5419   & 73.4314  & 0.0562   & 0.5550   \\
        & \textbf{MiniCPM-V-2-6} & 0.0001   & 0.2943   & 0.0604   & 6.2279   & -1.2576  & 0.0460   \\
        & \textbf{GPT-4o-mini} & 0.7975   & 79.7497  & 0.6845   & 89.2833  & \textbf{0.3699}   & 0.6830   \\ \midrule
        \multirow{6}{*}{\textbf{Flowchart}}
        & \textbf{Llama3.2-11B-Vision-Instruct} & 0.0004   & 0.7182   & 0.1278   & 22.1123  & -0.6942  & 0.2966   \\
        & \textbf{Llama-VL-TUG (ours)} & \textbf{0.9082} & \textbf{90.8177}  & \textbf{0.9724} & \textbf{97.5900} & \textbf{0.2340}  & \textbf{0.9864}   \\ 
        & \textbf{Gemma3-12B-Instruction-Tuned} & 0.0000   & 1.5966   & 0.1806   & 22.3908  & -0.6147  & 0.2900   \\
        & \textbf{Qwen2.5-VL-7B-Instruct} & 0.0094   & 2.3684   & 0.1568   & 23.2609  & -0.5725  & 0.2718   \\
        & \textbf{MiniCPM-V-2-6} & 0.0000   & 0.5603   & 0.0482   & 6.8664   & -1.1244  & 0.0517   \\
        & \textbf{GPT-4o-mini} & 0.0003   & 1.0921   & 0.0704   & 13.3751  & -1.0583  & 0.0437   \\ \midrule
        \multirow{6}{*}{\textbf{Graph}}
        & \textbf{Llama3.2-11B-Vision-Instruct} & 0.1080   & 14.7753  & 0.4323   & 62.3889  & \textbf{0.2716}   & 0.6220   \\
        & \textbf{Llama-VL-TUG (ours)} & \textbf{0.6360} & \textbf{63.8309}   & \textbf{0.7667} & \textbf{86.2771} & -0.0717 & \textbf{0.7047}   \\ 
        & \textbf{Gemma3-12B-Instruction-Tuned} & 0.1370   & 17.1570  & 0.4911   & 62.5730  & 0.0750   & 0.5705   \\
        & \textbf{Qwen2.5-VL-7B-Instruct} & 0.0861   & 12.3727  & 0.4571   & 56.7718  & -0.0477  & 0.5189   \\
        & \textbf{MiniCPM-V-2-6} & 0.0000   & 2.4965   & 0.1444   & 7.8484   & -1.2013  & 0.0544   \\
        & \textbf{GPT-4o-mini} & 0.1672   & 18.5799  & 0.4880   & 57.8477  & -0.0880  & 0.5753   \\ \midrule
        \multirow{7}{*}{\textbf{Packet}}
        & \textbf{Llama3.2-11B-Vision-Instruct} & 0.0298   & 3.2621   & 0.1502   & 24.7333  & -0.8368  & 0.2480   \\
        & \textbf{Llama-VL-TUG (ours)} & \textbf{0.5190} & \textbf{51.8954}   & \textbf{0.6775} & \textbf{68.7837} & \textbf{0.0244}  & \textbf{0.5898}   \\ 
        & \textbf{Gemma3-12B-Instruction-Tuned} & 0.0278   & 2.8232   & 0.0941   & 14.5225  & -1.2830  & 0.1566   \\
        & \textbf{Qwen2.5-VL-7B-Instruct} & 0.0450   & 4.8357   & 0.2034   & 29.4239  & -0.7568  & 0.2976   \\
        & \textbf{MiniCPM-V-2-6} & 0.0000   & 0.1241   & 0.0212   & 5.5846   & -1.2374  & 0.0230   \\
        & \textbf{GPT-4o-mini} & 0.0525   & 5.7991   & 0.2362   & 38.8445  & -0.3666  & 0.3684   \\ \midrule
        \multirow{6}{*}{\textbf{Sequence}}
        & \textbf{Llama3.2-11B-Vision-Instruct} & 0.2053   & 21.3545  & 0.5646   & 65.3644  & -0.1650  & 0.4853   \\
        & \textbf{Llama-VL-TUG (ours)} & \textbf{0.8236} & \textbf{82.3598}   & \textbf{0.9555} & \textbf{96.2541} & 0.0992  & \textbf{0.9606}   \\ 
        & \textbf{Gemma3-12B-Instruction-Tuned} & 0.2798   & 28.2498  & 0.6416   & 73.1667  & \textbf{0.1889}   & 0.5860   \\
        & \textbf{Qwen2.5-VL-7B-Instruct} & 0.3008   & 30.8903  & 0.6368   & 73.6891  & 0.1591   & 0.6002   \\
        & \textbf{MiniCPM-V-2-6} & 0.0000   & 1.7125   & 0.1946   & 23.0416  & -1.3419  & 0.0696   \\
        & \textbf{GPT-4o-mini} & 0.1448   & 18.9303  & 0.5875   & 67.1868  & -0.2169  & 0.4863   \\ \midrule
        \multirow{6}{*}{\textbf{State}}
        & \textbf{Llama3.2-11B-Vision-Instruct} & 0.4643   & 46.4642  & 0.4617   & 58.4349  & -0.0451  & 0.4639   \\
        & \textbf{Llama-VL-TUG (ours)} & \textbf{0.8793} & \textbf{87.9324}   & \textbf{0.7923} & \textbf{91.4462} & \textbf{0.4082}  & \textbf{0.6738}   \\
        & \textbf{Gemma3-12B-Instruction-Tuned} & 0.2088   & 21.3466  & 0.2781   & 40.4843  & -0.3614  & 0.4356   \\
        & \textbf{Qwen2.5-VL-7B-Instruct} & 0.2609   & 26.4496  & 0.3161   & 45.0595  & -0.3306  & 0.4507   \\
        & \textbf{MiniCPM-V-2-6} & 0.0003   & 0.3201   & 0.1123   & 3.5678   & -1.3960  & 0.0091   \\
        & \textbf{GPT-4o-mini} & 0.5100   & 51.0173  & 0.5167   & 61.4020  & 0.0565   & 0.5049   \\
        \bottomrule          
    \end{tabular}
}
    \caption{Detailed results for Image2Code Generation (On Synthetic Eval Dataset)}
    \label{tab:image_2_code_metrics}
\end{table*}

\begin{table*}[ht]
    \centering
    {\fontsize{7}{9}\selectfont

    \renewcommand{\arraystretch}{1.2}
    
    \begin{tabular}{llcccccc}
        \toprule
        \textbf{Diagram Type} & \textbf{Model} & \textbf{BLEU$\uparrow$} & \textbf{SACREBLEU$\uparrow$} & \textbf{METEOR$\uparrow$} & \textbf{chrF$\uparrow$} & \textbf{BLEURT$\uparrow$} & \textbf{ROUGE-L$\uparrow$} \\
        \midrule
        \multirow{6}{*}{\textbf{Block}}          
        & \textbf{Llama3.2-11B-Vision-Instruct} & 0.0290   & 4.2783   & 0.2111   & 32.7885  & -0.5828  & 0.2457   \\
        & \textbf{Llama-VL-TUG (ours)} & \textbf{0.8994 }& \textbf{89.9374}   & \textbf{0.9547} & \textbf{94.6979} & \textbf{0.7247} & \textbf{0.8440}   \\ 
        & \textbf{Gemma3-12B-Instruction-Tuned} & 0.0663   & 6.7988   & 0.2753   & 34.8009  & -0.5153  & 0.2684   \\
        & \textbf{Qwen2.5-VL-7B-Instruct} & 0.0301   & 5.1130   & 0.2592   & 27.9851  & -0.6851  & 0.2418   \\
        & \textbf{MiniCPM-V-2-6} & 0.0168   & 3.5740   & 0.2355   & 22.2868  & -0.7862  & 0.2055   \\
        & \textbf{GPT-4o-mini} & 0.1080   & 10.9790  & 0.2684   & 30.7947  & -0.5459  & 0.2426   \\ \midrule
        \multirow{6}{*}{\textbf{C4}}
        & \textbf{Llama3.2-11B-Vision-Instruct} & 0.0008   & 0.7302   & 0.0965   & 23.2782  & -0.6560  & 0.2463   \\
        & \textbf{Llama-VL-TUG (ours)} & \textbf{0.9609} & \textbf{96.0880}   & \textbf{0.9824} & \textbf{97.8606} & \textbf{0.4821} & \textbf{0.9711}   \\ 
        & \textbf{Gemma3-12B-Instruction-Tuned} & 0.0093   & 0.9878   & 0.1259   & 23.6268  & -0.6678  & 0.2728   \\
        & \textbf{Qwen2.5-VL-7B-Instruct} & 0.0059   & 1.2289   & 0.1066   & 24.0475  & -0.6474  & 0.2711   \\
        & \textbf{MiniCPM-V-2-6} & 0.0129   & 1.9129   & 0.1401   & 21.1324  & -0.7070  & 0.2614   \\
        & \textbf{GPT-4o-mini} & 0.8229   & 82.2867  & 0.8841   & 88.4546  & 0.2825   & 0.8299   \\ \midrule
        \multirow{6}{*}{\textbf{Class}}
        & \textbf{Llama3.2-11B-Vision-Instruct} & 0.3238   & 32.3804  & 0.4820   & 60.4781  & -0.3297  & 0.6389   \\
        & \textbf{Llama-VL-TUG (ours)} & \textbf{0.9656} & \textbf{96.5596}   & \textbf{0.9889} & \textbf{98.5066} & \textbf{0.6291} & \textbf{0.9789}   \\ 
        & \textbf{Gemma3-12B-Instruction-Tuned} & 0.3469   & 34.7061  & 0.3975   & 54.5017  & -0.4732  & 0.5856   \\
        & \textbf{Qwen2.5-VL-7B-Instruct} & 0.3491   & 34.9999  & 0.4559   & 61.5165  & -0.3032  & 0.6486   \\
        & \textbf{MiniCPM-V-2-6} & 0.2144   & 21.7403  & 0.3136   & 51.1385  & -0.3994  & 0.5044   \\
        & \textbf{GPT-4o-mini} & 0.9029   & 90.2938  & 0.9571   & 94.9272  & 0.5473   & 0.9547   \\ \midrule
        \multirow{6}{*}{\textbf{Flowchart}}
        & \textbf{Llama3.2-11B-Vision-Instruct} & 0.0014   & 1.2677   & 0.1303   & 22.7094  & -0.5936  & 0.2582   \\
        & \textbf{Llama-VL-TUG (ours)} & \textbf{0.9656} & 9\textbf{6.5596}    & \textbf{0.9889} & \textbf{98.5066} & \textbf{0.6291} & \textbf{0.9789}   \\ 
        & \textbf{Gemma3-12B-Instruction-Tuned} & 0.0204   & 2.6764   & 0.1725   & 24.7587  & -0.6932  & 0.2767   \\
        & \textbf{Qwen2.5-VL-7B-Instruct} & 0.0067   & 2.0388   & 0.1526   & 22.7352  & -0.5851  & 0.2788   \\
        & \textbf{MiniCPM-V-2-6} & 0.0002   & 1.5939   & 0.1540   & 21.7691  & -0.5673  & 0.2961   \\
        & \textbf{GPT-4o-mini} & 0.1756   & 17.5553  & 0.4283   & 29.6297  & -1.0116  & 0.1797   \\ \midrule
        \multirow{6}{*}{\textbf{Graph}}
        & \textbf{Llama3.2-11B-Vision-Instruct} & 0.1428   & 17.8969  & 0.5116   & 66.2555  & 0.1393   & 0.6608   \\
        & \textbf{Llama-VL-TUG (ours)} & 0.8464 & 84.6445   & 0.8541 & 93.3428 & 0.7006 & 0.9403   \\ 
        & \textbf{Gemma3-12B-Instruction-Tuned} & 0.1590   & 20.8206  & 0.5247   & 73.1211  & 0.2322   & 0.7391   \\
        & \textbf{Qwen2.5-VL-7B-Instruct} & 0.0563   & 9.7717   & 0.4591   & 57.6757  & -0.0484  & 0.5707   \\
        & \textbf{MiniCPM-V-2-6} & 0.0115   & 5.3132   & 0.3373   & 50.4762  & -0.3247  & 0.4465   \\
        & \textbf{GPT-4o-mini} & \textbf{0.9092}   & \textbf{90.9247}  & \textbf{0.9076}   & \textbf{97.3794}  & \textbf{0.9716}   & \textbf{0.9990}   \\ 
        \midrule
        \multirow{6}{*}{\textbf{Packet}}
        & \textbf{Llama3.2-11B-Vision-Instruct} & 0.0235   & 3.6337   & 0.1563   & 27.1367  & -0.8113  & 0.2517   \\
        & \textbf{Llama-VL-TUG (ours)} & 0.9580 & 95.8048   & 0.9902 & 97.0555 & 0.8043 & 0.9803   \\ 
        & \textbf{Gemma3-12B-Instruction-Tuned} & 0.1243   & 12.7378  & 0.3533   & 45.9207  & -0.3478  & 0.4548   \\
        & \textbf{Qwen2.5-VL-7B-Instruct} & 0.0544   & 5.7997   & 0.2202   & 27.7765  & -0.8272  & 0.2582   \\
        & \textbf{MiniCPM-V-2-6} & 0.0556   & 5.8855   & 0.2378   & 34.0349  & -0.5503  & 0.3326   \\
        & \textbf{GPT-4o-mini} & \textbf{1.0000}   & \textbf{100.0000} & \textbf{1.0000}   & \textbf{100.0000} & \textbf{0.8295}   & \textbf{1.0000}   \\ \midrule
        \multirow{6}{*}{\textbf{Sequence}}
        & \textbf{Llama3.2-11B-Vision-Instruct} & 0.2793   & 28.1335  & 0.6721   & 74.8698  & -0.3530  & 0.5829   \\
        & \textbf{Llama-VL-TUG (ours)} & \textbf{1.0000} & \textbf{100.0000} & \textbf{1.0000} & \textbf{100.0000} & \textbf{1.0068} & \textbf{1.0000}   \\
        & \textbf{Gemma3-12B-Instruction-Tuned} & 0.4614   & 46.2925  & 0.7616   & 88.6449  & 0.2405   & 0.8791   \\
        & \textbf{Qwen2.5-VL-7B-Instruct} & 0.3698   & 37.6440  & 0.6798   & 80.9233  & 0.1786   & 0.7054   \\
        & \textbf{MiniCPM-V-2-6} & 0.2580   & 26.6537  & 0.5595   & 72.8661  & -0.1069  & 0.5815   \\
        & \textbf{GPT-4o-mini} & 0.8511   & 85.1115  & 0.8815   & 96.7221  & 0.8917   & 0.9997   \\ \midrule
        \multirow{6}{*}{\textbf{State}}
        & \textbf{Llama3.2-11B   } & 0.2885   & 28.8777  & 0.2065   & 47.7941  & -0.4619  & 0.4204   \\
        & \textbf{Llama-VL-TUG (ours)} & \textbf{0.9369} & \textbf{93.6941}   & \textbf{0.9380} & \textbf{95.9338} & \textbf{0.7330} & \textbf{0.9699}   \\
        & \textbf{Gemma3-12B-Instruction-Tuned} & 0.5186   & 51.8634  & 0.3189   & 79.6331  & 0.2222   & 0.8198   \\
        & \textbf{Qwen2.5-VL-7B-Instruct} & 0.3800   & 38.0818  & 0.2994   & 58.7889  & -0.2469  & 0.5391   \\
        & \textbf{MiniCPM-V-2-6} & 0.2960   & 29.6741  & 0.2650   & 50.5644  & -0.4541  & 0.4362   \\
        & \textbf{GPT-4o-mini} & 0.9278   & 92.7769  & 0.9365   & 94.8294  & 0.7213   & 0.9582   \\
        \bottomrule
    \end{tabular}
}
    \caption{Detailed results for Description2Code Generation (On Synthetic Eval Dataset)}
    \label{tab:summary_2_code_metrics}
\end{table*}

\begin{table*}[ht]
    \centering
    \renewcommand{\arraystretch}{1.2}
    {\fontsize{7}{9}\selectfont % This sets font size to 7pt with a 9pt line spacing
    \begin{tabular}{llcccccc}
        \toprule
        \textbf{Diagram Type} & \textbf{Model} & \textbf{BLEU$\uparrow$} & \textbf{SACREBLEU$\uparrow$} & \textbf{METEOR$\uparrow$} & \textbf{chrF$\uparrow$} & \textbf{BLEURT$\uparrow$} & \textbf{ROUGE-L$\uparrow$}  \\
        \midrule
        \multirow{6}{*}{\textbf{Block}}          
        & \textbf{Llama3.2-11B-Vision-Instruct} & 0.6919   & 69.1903  & 0.8296   & 82.1518  & 0.3238   & 0.7618   \\
        & \textbf{Llama-VL-TUG (ours)} & \textbf{0.8952}  & \textbf{89.5213} & \textbf{0.9591}  & \textbf{97.7836} & \textbf{0.7190}  & \textbf{0.9561}  \\ 
        & \textbf{Gemma3-12B-Instruction-Tuned} & 0.7717   & 77.1740  & 0.8765   & 88.7206  & 0.4529   & 0.8000   \\
        & \textbf{Qwen2.5-VL-7B-Instruct} & 0.6644   & 66.4355  & 0.7662   & 80.0331  & 0.3056   & 0.7424   \\
        & \textbf{MiniCPM-V-2-6} & 0.3366   & 33.6623  & 0.4848   & 45.9604  & -0.8165  & 0.5542   \\
        & \textbf{GPT-4o-mini} & 0.6916   & 69.1613  & 0.8112   & 83.1176  & 0.4294   & 0.7540   \\ \midrule
        \multirow{6}{*}{\textbf{C4}}
        & \textbf{Llama3.2-11B-Vision-Instruct} & 0.5537   & 55.3658  & 0.7376   & 73.4240  & 0.1010   & 0.6860   \\
        & \textbf{Llama-VL-TUG (ours)} & \textbf{0.9916} & \textbf{99.1557} & \textbf{0.9993} & \textbf{99.9638} & \textbf{0.7393} & \textbf{1.0000}   \\ 
        & \textbf{Gemma3-12B-Instruction-Tuned} & 0.7599   & 75.9947  & 0.8790   & 87.2885  & 0.4770   & 0.8213   \\
        & \textbf{Qwen2.5-VL-7B-Instruct} & 0.7079   & 70.7854  & 0.8340   & 84.7851  & 0.4182   & 0.8037   \\
        & \textbf{MiniCPM-V-2-6} & 0.1583   & 15.8265  & 0.3620   & 36.0967  & -0.4531  & 0.4934   \\
        & \textbf{GPT-4o-mini} & 0.8198   & 81.9774  & 0.9085   & 90.8801  & 0.5639   & 0.8511   \\ \midrule
        \multirow{6}{*}{\textbf{Class}}
        & \textbf{Llama3.2-11B-Vision-Instruct} & 0.3256   & 32.5603  & 0.4865   & 50.6936  & -0.2765  & 0.5287   \\
        & \textbf{Llama-VL-TUG (ours)} & \textbf{0.9746} & \textbf{97.4633} & \textbf{0.9809} & \textbf{98.0992} & \textbf{0.6531} & \textbf{0.9764}   \\ 
        & \textbf{Gemma3-12B-Instruction-Tuned} & 0.6978   & 69.7829  & 0.6972   & 77.1858  & 0.2685   & 0.6276   \\
        & \textbf{Qwen2.5-VL-7B-Instruct} & 0.7294   & 72.9380  & 0.7439   & 82.1311  & 0.3192   & 0.6571   \\
        & \textbf{MiniCPM-V-2-6} & 0.3256   & 32.5603  & 0.4865   & 50.6936  & -0.2765  & 0.5287   \\
        & \textbf{GPT-4o-mini} & 0.8032   & 80.3173  & 0.8333   & 87.2117  & 0.3869   & 0.6836   \\ \midrule
        \multirow{6}{*}{\textbf{Flowchart}}
        & \textbf{Llama3.2-11B-Vision-Instruct} & 0.6598   & 65.9797  & 0.7943   & 82.5270  & 0.3285   & 0.8450   \\
        & \textbf{Llama-VL-TUG (ours)} & \textbf{0.9914} & \textbf{99.1419} & \textbf{0.9993} & \textbf{99.9588} & \textbf{0.8069} & \textbf{1.0000}  \\ 
        & \textbf{Gemma3-12B-Instruction-Tuned} & 0.4695   & 46.9527  & 0.6673   & 73.1234  & 0.2282   & 0.9125   \\
        & \textbf{Qwen2.5-VL-7B-Instruct} & 0.6823   & 68.2324  & 0.8161   & 84.3189  & 0.4034   & 0.8936   \\
        & \textbf{MiniCPM-V-2-6} & 0.2970   & 29.7045  & 0.5195   & 47.7263  & -0.9542  & 0.4732   \\
        & \textbf{GPT-4o-mini} & 0.2728   & 27.2827  & 0.5799   & 50.0034  & -0.9770  & 0.4993   \\ \midrule
        \multirow{6}{*}{\textbf{Graph}}
        & \textbf{Llama3.2-11B-Vision-Instruct} & 0.5314   & 53.1421  & 0.7467   & 73.3913  & 0.2576   & 0.6943   \\
        & \textbf{Llama-VL-TUG (ours)} & \textbf{0.8878} & \textbf{88.7774} & \textbf{0.9654} & \textbf{95.9720} & \textbf{0.7278} & \textbf{0.9436}   \\ 
        & \textbf{Gemma3-12B-Instruction-Tuned} & 0.5901   & 59.0075  & 0.7873   & 77.8164  & 0.3555   & 0.7381   \\
        & \textbf{Qwen2.5-VL-7B-Instruct} & 0.5296   & 52.9603  & 0.7175   & 73.5296  & 0.2960   & 0.7127   \\
        & \textbf{MiniCPM-V-2-6} & 0.5314   & 53.1421  & 0.7467   & 73.3913  & 0.2576   & 0.6943   \\
        & \textbf{GPT-4o-mini} & 0.5609   & 56.0883  & 0.7868   & 76.2206  & 0.2698   & 0.7030   \\ \midrule
        \multirow{6}{*}{\textbf{Packet}}
        & \textbf{Llama3.2-11B-Vision-Instruct} & 0.4044  & 40.4410 & 0.6921  & 74.9078 & 0.2912  & 0.5578   \\
        & \textbf{Llama-VL-TUG (ours)} & \textbf{0.6302}  & \textbf{63.0240} & \textbf{0.7891}  & \textbf{81.7360} & \textbf{0.4825}  & \textbf{0.7232}   \\ 
        & \textbf{Gemma3-12B-Instruction-Tuned} & 0.4553.  & 45.5292  & 0.6432   & 70.5237  & 0.2239   & 0.6365   \\
        & \textbf{Qwen2.5-VL-7B-Instruct} & 0.3441   & 34.4130  & 0.6457   & 71.2078  & 0.1079   & 0.5171   \\
        & \textbf{MiniCPM-V-2-6} & 0.4044   & 40.4410  & 0.6921   & 74.9078  & 0.2912   & 0.5578   \\
        & \textbf{GPT-4o-mini} & 0.3828   & 38.2805  & 0.6982   & 75.0080  & 0.1336   & 0.5470   \\
        \midrule
        \multirow{6}{*}{\textbf{Sequence}}
        & \textbf{Llama3.2-11B-Vision-Instruct} & 0.6893   & 68.9273  & 0.8780   & 84.3601  & 0.3644   & 0.7752   \\
        & \textbf{Llama-VL-TUG (ours)} & \textbf{0.8888}  & \textbf{88.8756} & \textbf{0.9775}  & \textbf{97.2891} & \textbf{0.6180}  & \textbf{0.9603}   \\ 
        & \textbf{Gemma3-12B-Instruction-Tuned} & 0.5279   & 52.7857  & 0.7633   & 79.9958  & 0.2875   & 0.7664   \\
        & \textbf{Qwen2.5-VL-7B-Instruct} & 0.6505   & 65.0489  & 0.8374   & 82.4184  & 0.3669   & 0.7782   \\
        & \textbf{MiniCPM-V-2-6} & 0.6893   & 68.9273  & 0.8780   & 84.3601  & 0.3644   & 0.7752   \\
        & \textbf{GPT-4o-mini} & 0.7568   & 75.6781  & 0.9242   & 92.3463  & 0.5460   & 0.8144   \\ \midrule
        \multirow{6}{*}{\textbf{State}}
        & \textbf{Llama3.2-11B-Vision-Instruct} & 0.6220   & 62.1973  & 0.7418   & 75.4259  & 0.3177   & 0.6388   \\
        & \textbf{Llama-VL-TUG (ours)} & \textbf{0.8919}  & \textbf{89.1866} & \textbf{0.9227}  & \textbf{95.2614} & \textbf{0.6111}  & \textbf{0.8635}   \\
        & \textbf{Gemma3-12B-Instruction-Tuned} & 0.6292   & 62.9213  & 0.7348   & 75.3195  & 0.4053   & 0.6640   \\
        & \textbf{Qwen2.5-VL-7B-Instruct} & 0.6082   & 60.8156  & 0.6896   & 76.0601  & 0.3481   & 0.6437   \\
        & \textbf{MiniCPM-V-2-6} & 0.1269   & 12.6916  & 0.4379   & 28.6882  & -0.8319  & 0.3387   \\
        & \textbf{GPT-4o-mini} & 0.6634   & 66.3353  & 0.7468   & 79.3858  & 0.4126   & 0.6664   \\
        \bottomrule
    \end{tabular}
    }
    \caption{Detailed results for Image2Description (On Synthetic Eval Dataset)}
    \label{tab:topological_summary_metrics}
\end{table*}

\begin{table*}[ht]
    \centering
    \renewcommand{\arraystretch}{1.2}
    {\fontsize{7}{9}\selectfont % This sets font size to 7pt with a 9pt line spacing
    \begin{tabular}{llcccccc}
        \toprule

\textbf{Diagram Type} & \textbf{Model} & \textbf{BLEU$\uparrow$} & \textbf{SACREBLEU$\uparrow$} & \textbf{METEOR$\uparrow$} & \textbf{chrF$\uparrow$} & \textbf{BLEURT$\uparrow$} & \textbf{ROUGE-L$\uparrow$} \\
\midrule
\multirow{7}{*}{\textbf{Block}} 
& \textbf{Llama3.2-11B-Vision-Instruct} & 0.0211 & 4.0663 & 0.2446 & 28.1006 & -0.8078 & 0.2623 \\
& \textbf{Gemma3-12B-Instruction-Tuned} & 0.0219 & 4.8489 & 0.2844 & 21.8190 & -0.8729 & 0.2373 \\
& \textbf{Qwen2.5-VL-7B-Instruct} & 0.0200 & 4.8255 & 0.2931 & 22.1637 & -0.8625 & 0.2505 \\
& \textbf{MiniCPM-V-2-6} & 0.0303 & 4.9937 & 0.3025 & 32.6082 & -0.7513 & 0.1926 \\
& \textbf{GPT-4o-mini} & 0.0044 & 3.5151 & 0.2701 & 19.9548 & -0.8435 & 0.2395 \\
& \textbf{Llama-VL-TUG (ours)} & \textbf{0.4905} & \textbf{49.1096} & \textbf{0.7208} & \textbf{72.1252} & \textbf{-0.1484} & \textbf{0.6163} \\
& \textbf{Llama3.2-w/o Self Supervision (Ablation)} & 0.4089 & 41.5985 & 0.6412 & 64.3320 & -0.2587 & 0.5343 \\
\midrule
\multirow{7}{*}{\textbf{C4}} 
& \textbf{Llama3.2-11B-Vision-Instruct} & 0.0261 & 3.4245 & 0.1522 & 31.3062 & -0.7199 & 0.3490 \\
& \textbf{Gemma3-12B-Instruction-Tuned} & 0.0814 & 8.6327 & 0.2431 & 35.5166 & -0.7242 & 0.3587 \\
& \textbf{Qwen2.5-VL-7B-Instruct} & 0.0338 & 4.3824 & 0.1820 & 31.1980 & -0.7247 & 0.3519 \\
& \textbf{MiniCPM-V-2-6} & 0.0471 & 5.2783 & 0.2358 & 36.5629 & -0.6841 & 0.2360 \\
& \textbf{GPT-4o-mini} & 0.0363 & 4.4410 & 0.1453 & 29.5557 & \textbf{-0.7419} & 0.3685 \\
& \textbf{Llama-VL-TUG (ours)} & \textbf{0.4893} & \textbf{48.9480} & \textbf{0.6438} & 65.7211 & -0.2596 & 0.5371 \\
& \textbf{Llama3.2-w/o Self Supervision (Ablation)} & 0.4731 & 47.4111 & 0.6185 & \textbf{66.3420} & -0.2626 & \textbf{0.5412} \\
\midrule
\multirow{7}{*}{\textbf{Class}} 
& \textbf{Llama3.2-11B-Vision-Instruct} & 0.4109 & 41.4988 & 0.5741 & 68.3856 & -0.2471 & 0.7046 \\
& \textbf{Gemma3-12B-Instruction-Tuned} & 0.3773 & 38.3370 & 0.5416 & 68.4976 & -0.2581 & 0.6865 \\
& \textbf{Qwen2.5-VL-7B-Instruct} & 0.3402 & 34.3344 & 0.5597 & 66.7691 & -0.3037 & 0.6625 \\
& \textbf{MiniCPM-V-2-6} & 0.0606 & 6.8319 & 0.2751 & 40.6963 & -0.6234 & 0.2985 \\
& \textbf{GPT-4o-mini} & \textbf{0.5313} & \textbf{53.2315} & \textbf{0.7255} & \textbf{80.8516} & \textbf{-0.0819} & \textbf{0.8068} \\
& \textbf{Llama-VL-TUG (ours)} & 0.3378 & 34.1778 & 0.5708 & 62.9082 & -0.3636 & 0.5375 \\
& \textbf{Llama3.2-w/o Self Supervision (Ablation)} & 0.3882 & 38.9145 & 0.6119 & 68.3929 & -0.3260 & 0.5932 \\
\midrule
\multirow{7}{*}{\textbf{Flowchart}} 
& \textbf{Llama3.2-11B-Vision-Instruct} & 0.0220 & 4.6773 & 0.2696 & 30.7875 & -0.6497 & 0.4238 \\
& \textbf{Gemma3-12B-Instruction-Tuned} & 0.2980 & 29.8013 & 0.6134 & 46.4670 & -0.1988 & 0.7449 \\
& \textbf{Qwen2.5-VL-7B-Instruct} & 0.1435 & 16.0567 & 0.4860 & 36.3134 & -0.4691 & 0.5332 \\
& \textbf{MiniCPM-V-2-6} & 0.1602 & 16.4982 & 0.4953 & 36.2944 & -0.4702 & 0.4470 \\
& \textbf{GPT-4o-mini} & \textbf{0.3241} & \textbf{32.4087} & \textbf{0.6488} & \textbf{51.1517} & \textbf{-0.1293} & \textbf{0.7786} \\
& \textbf{Llama-VL-TUG (ours)} & 0.0491 & 7.2717 & 0.3740 & 37.6756 & -0.7727 & 0.3009 \\
& \textbf{Llama3.2-w/o Self Supervision (Ablation)} & 0.0475 & 7.2187 & 0.3743 & 37.8416 & -0.7709 & 0.2990 \\
\midrule
\multirow{7}{*}{\textbf{Graph}} 
& \textbf{Llama3.2-11B-Vision-Instruct} & 0.0027 & 3.4573 & 0.2706 & 37.3223 & -0.6903 & 0.5086 \\
& \textbf{Gemma3-12B-Instruction-Tuned} & \textbf{0.0330} & 7.5894 & 0.3823 & 40.5827 & -0.5487 & 0.5809 \\
& \textbf{Qwen2.5-VL-7B-Instruct} & 0.0276 & \textbf{7.8834} & \textbf{0.4031} & 39.7404 & -0.5507 & 0.6109 \\
& \textbf{MiniCPM-V-2-6} & 0.0124 & 4.8452 & 0.3616 & 31.6561 & -0.6496 & 0.3859 \\
& \textbf{GPT-4o-mini} & 0.0223 & 6.8887 & 0.3710 & \textbf{40.6762} & \textbf{-0.5060} & \textbf{0.6607} \\
& \textbf{Llama-VL-TUG (ours)} & 0.0047 & 3.8026 & 0.2654 & 34.4467 & -0.6424 & 0.4685 \\
& \textbf{Llama3.2-w/o Self Supervision (Ablation)} & 0.0021 & 3.5804 & 0.2838 & 35.2435 & -0.6092 & 0.4621 \\
\midrule
\multirow{7}{*}{\textbf{Packet}} 
& \textbf{Llama3.2-11B-Vision-Instruct} & 0.0072 & 2.9514 & 0.2416 & 36.5142 & -0.7099 & 0.3938 \\
& \textbf{Gemma3-12B-Instruction-Tuned} & 0.0137 & 3.5934 & 0.2763 & 38.3844 & -0.6463 & 0.4247 \\
& \textbf{Qwen2.5-VL-7B-Instruct} & 0.0170 & 3.9062 & 0.2751 & 35.2018 & -0.7856 & 0.3460 \\
& \textbf{MiniCPM-V-2-6} & 0.0113 & 2.3418 & 0.2312 & 29.6449 & -0.7706 & 0.2127 \\
& \textbf{GPT-4o-mini} & 0.0071 & 3.4682 & 0.2533 & 39.6114 & -0.6467 & 0.4199 \\
& \textbf{Llama-VL-TUG (ours)} & \textbf{0.3479} & \textbf{34.7864} & \textbf{0.6074} & \textbf{60.0923} & \textbf{-0.3605} & \textbf{0.4838} \\
& \textbf{Llama3.2-w/o Self Supervision (Ablation)} & 0.3256 & 32.5567 & 0.5945 & 58.5569 & -0.3824 & 0.4566 \\
\midrule
\multirow{7}{*}{\textbf{Sequence}} 
& \textbf{Llama3.2-11B-Vision-Instruct} & 0.2644 & 27.9428 & 0.5901 & 66.7506 & -0.3793 & 0.6272 \\
& \textbf{Gemma3-12B-Instruction-Tuned} & 0.3429 & 35.1075 & 0.7068 & 76.7020 & \textbf{-0.2194} & \textbf{0.7590} \\
& \textbf{Qwen2.5-VL-7B-Instruct} & 0.2935 & 30.4405 & 0.6523 & 73.3296 & -0.3141 & 0.6859 \\
& \textbf{MiniCPM-V-2-6} & 0.1913 & 20.5521 & 0.5234 & 59.4344 & -0.4340 & 0.5117 \\
& \textbf{GPT-4o-mini} & \textbf{0.4261} & \textbf{43.8398} & \textbf{0.7140} & \textbf{77.8157} & -0.2344 & 0.7504 \\
& \textbf{Llama-VL-TUG (ours)} & 0.1400 & 14.9873 & 0.4974 & 53.9343 & -0.4537 & 0.5105 \\
& \textbf{Llama3.2-w/o Self Supervision (Ablation)} & 0.1275 & 14.1303 & 0.5016 & 53.7913 & \textbf{-0.2194} & 0.5196 \\
\midrule
\multirow{7}{*}{\textbf{State}} 
& \textbf{Llama3.2-11B-Vision-Instruct} & 0.1187 & 14.5587 & 0.4231 & 50.3392 & -0.5306 & 0.4635 \\
& \textbf{Gemma3-12B-Instruction-Tuned} & 0.2861 & 29.9928 & 0.6579 & 61.0823 & -0.4059 & 0.5660 \\
& \textbf{Qwen2.5-VL-7B-Instruct} & 0.0922 & 12.3810 & 0.4776 & 46.7963 & -0.5155 & 0.4990 \\
& \textbf{MiniCPM-V-2-6} & 0.0195 & 4.4052 & 0.3851 & 32.9893 & -0.7931 & 0.2421 \\
& \textbf{GPT-4o-mini} & \textbf{0.3910} & \textbf{39.4981} & \textbf{0.7483} & \textbf{71.9773} & \textbf{-0.2527} & \textbf{0.6688} \\
& \textbf{Llama-VL-TUG (ours)} & 0.0510 & 6.5729 & 0.3668 & 34.2009 & -0.9271 & 0.1671 \\
& \textbf{Llama3.2-w/o Self Supervision (Ablation)} & 0.0200 & 3.2555 & 0.2795 & 25.6581 & -0.9875 & 0.1300 \\
\bottomrule
\end{tabular}
}

\caption{Detailed results for Image2Code (On Real-world Corpus \textbf{D3})}
\label{tab:auto_real_world}
\end{table*}

\begin{table*}[ht]
    \centering
    \small
    \renewcommand{\arraystretch}{1.2}
    \begin{adjustbox}{width=\textwidth}
    \begin{tabular}{llccccccc}
        \toprule
        \textbf{Diagram Type} & \textbf{Model} &\textbf{BLEU$\uparrow$} & \textbf{SACREBLEU$\uparrow$} & \textbf{CodeBLEU$\uparrow$} & \textbf{METEOR$\uparrow$} & \textbf{chrF$\uparrow$} & \textbf{BLEURT$\uparrow$} & \textbf{ROUGE-L$\uparrow$} \\
        \midrule
        \multirow{7}{*}{\textbf{C4}}          
        & Llama3.2-11B-Vision-Instruct &   0.0649	 & 6.7359	 & 0.0649 &	0.1856	& 42.4385 &	-0.4850 & 0.4000  \\
        & Qwen2.5-VL-7B-Instruct   &  0.0297 &	3.6166 &	0.0297 &	0.1497 &	33.6653 &	-0.6413 &	0.3342  \\
        & MiniCPM-V-2-6     & 0.0000 & 0.0439 & 0.0000 & 0.0149 & 3.1508 & -1.2448 & 0.0517 \\
        & GPT-4o-mini  & 0.1022 &	10.4743 & 0.1022 &	0.2325 &	44.4395 &	-0.4793 &	0.4897  \\
        & \textbf{Llama3.2-11B-Vis-Inst-finetuned (CTS1)}&0.9991 & 99.9065 & 0.9991 & 0.9997 & 99.9171 & 0.5515 & 0.9977 \\
        & \textbf{Llama3.2-11B-Vis-Inst-finetuned (CTS1 + CTS2)}&0.9994 & 99.9426 & 0.9994 & 0.9997 & 99.9373 & \textbf{0.5535} & 0.9988 \\
        & \textbf{Llama3.2-11B-Vis-Inst-finetuned (CTS1 + CTS2 + CTS3)}&\textbf{0.9996} & \textbf{99.9664} &\textbf{ 0.9996} &\textbf{ 0.9998} & \textbf{99.9646} & 0.5534 & \textbf{0.9990} \\ \midrule
        
        \multirow{7}{*}{\textbf{Flowchart}}
        & Llama3.2-11B-Vision-Instruct   &  0.1614	& 16.1660 &	0.1614 &	0.3073 &	48.9088	& -0.5158 &	0.3095  \\
        & Qwen2.5-VL-7B-Instruct   &  0.1620 &	16.2131 &	0.1620 &	0.3482 &	44.8345 &	-0.4934 &	0.3666  \\
        & MiniCPM-V-2-6     & 0.0006 & 0.4957 & 0.0006 & 0.0501 & 6.0572 & -1.1414 & 0.0798  \\
        & GPT-4o-mini  & 0.1632 &	16.3373 &	0.1632 &	0.3500 &	46.7228 &	-0.2477 &	0.3884 \\
        & \textbf{Llama3.2-11B-Vis-Inst-finetuned (CTS1)}&\textbf{0.8844} & \textbf{88.4441} & \textbf{0.8844} & \textbf{0.9451} & \textbf{94.3313} & \textbf{0.6049} & 0.7234 \\
        & \textbf{Llama3.2-11B-Vis-Inst-finetuned (CTS1 + CTS2)}&0.8823 & 88.2370 & 0.8823 & 0.9429 &94.1437 & 0.5934 & 0.7329 \\
        & \textbf{Llama3.2-11B-Vis-Inst-finetuned (CTS1 + CTS2 + CTS3)}&0.8824 & 88.2425 & 0.8824 & 0.9447 & 94.3292 & 0.5918 & \textbf{0.7358} \\ \midrule
        
        \multirow{7}{*}{\textbf{State}}
        & Llama3.2-11B-Vision-Instruct  &   0.5589 &	55.9227 &	0.5589 &	0.5546 &	66.3656 &	-0.0411 &	0.4999  \\
        & Qwen2.5-VL-7B-Instruct   &  0.2674 &	26.9351 &	0.2674 &	0.3504 &	41.8751 &	-0.5321 &	0.4237  \\
        & MiniCPM-V-2-6     & 0.0000 & 0.2293 & 0.0000 & 0.0484 & 2.5240 & -1.4924 & 0.0053  \\
        & GPT-4o-mini &   0.5099 &	51.0172 &	0.5099 &	0.5167 &	61.4019 &	0.0564 &	0.5048  \\
        & \textbf{Llama3.2-11B-Vis-Inst-Finetuned (CTS1)} & \textbf{0.9416} & \textbf{94.1600} & \textbf{0.9416 }& \textbf{0.8983} & \textbf{95.4549} & 0.7688 & \textbf{0.6952} \\
        & \textbf{Llama3.2-11B-Vis-Inst-Finetuned (CTS1 + CTS2)} & 0.9358 & 93.5819 & 0.9358 & 0.8939 & 95.1734 & 0.7651 & 0.6911 \\
        & \textbf{Llama3.2-11B-Vis-Inst-Finetuned (CTS1 + CTS2 + CTS3)} &0.9387 & 93.8797 & 0.9387 & 0.8951 & 95.2143 & \textbf{0.7700} & 0.6950 \\ \midrule
        
        \multirow{7}{*}{\textbf{Block}}
        & Llama3.2-11B-Vision-Instruct &    0.0068 &	2.5596 &	0.0068 &	0.2026 &	29.6349 &	-0.7379 &	0.2323  \\
        & Qwen2.5-VL-7B-Instruct   &  0.0138 &	3.0087 &	0.0138 &	0.2116 &	25.0251 &	-0.8202 &	0.2265 \\
        & MiniCPM-V-2-6     & 0.0000 & 0.5429 & 0.0000 & 0.0765 & 5.1334 & -1.3276 & 0.0182 \\
        & GPT-4o-mini &    0.0020 &	2.4243 &	0.0020 &	0.1959 &	18.9215 &	-0.7491 &	0.2076 \\
        & \textbf{Llama3.2-11B-Vis-Inst-finetuned (CTS1)}& \textbf{0.9473} & \textbf{94.7300} & \textbf{0.9473} & 0.9885 & \textbf{97.9700} & 0.7667 & \textbf{0.9157} \\
        & \textbf{Llama3.2-11B-Vis-Inst-finetuned (CTS1 + CTS2)}&0.9460 & 94.6066 & 0.9460 & \textbf{0.9863} & 97.8143 & 0.7665 & \textbf{0.9157} \\
        & \textbf{Llama3.2-11B-Vis-Inst-finetuned (CTS1 + CTS2 + CTS3)}&0.9455 & 94.5507 & 0.9455 & 0.9882 & 97.8689 & \textbf{0.7668} & 0.9145 \\ \midrule
        
        \multirow{7}{*}{\textbf{Sequence}}
        & Llama3.2-11B-Vision-Instruct &     0.1852 &	19.4537 &	0.1852 &	0.5470 &	64.0124 &	-0.4667 &	0.4887  \\
        & Qwen2.5-VL-7B-Instruct  &   0.2851 &	29.0668 &	0.2851 &	0.6211 &	72.1925 &	-0.3264 &	0.6055  \\
        & MiniCPM-V-2-6     & 0.0000 & 1.1295 & 0.0000 & 0.0704 & 5.3291 & -1.4662 & 0.0037  \\
        & GPT-4o-mini &  0.1447 &	18.9302 &	0.1447 &	0.5875 &	67.1867 &	-0.2169 &	0.4862 \\
        & \textbf{Llama3.2-11B-Vis-Inst-finetuned (CTS1)}& \textbf{0.9974} & \textbf{99.7400} & \textbf{0.9974} & \textbf{0.9990} & \textbf{99.8600} & \textbf{1.0060} & \textbf{0.9981} \\
        & \textbf{Llama3.2-11B-Vis-Inst-finetuned (CTS1 + CTS2)}&0.9945 & 99.4514 & 0.9945 & 0.9978 & 99.6812 & 1.0045 & 0.9959 \\
        & \textbf{Llama3.2-11B-Vis-Inst-finetuned (CTS1 + CTS2 + CTS3)}&0.9944 & 99.4404 & 0.9944 & 0.9976 & 99.6791 & 1.0034 & 0.9957 \\ \midrule
        
        \multirow{7}{*}{\textbf{Graph}}
        & Llama3.2-11B-Vision-Instruct   & 0.1312 &	16.2506 &	0.1312 &	0.4702 &	63.0973	& -0.1654 &	0.6182   \\
        & Qwen2.5-VL-7B-Instruct  &  0.1078 &	13.6070 &	0.1078 &	0.4964 &	56.9693 &	-0.3456 &	0.5153  \\
        &  MiniCPM-V-2-6    & 0.0000 & 2.0048 & 0.0000 & 0.1186 & 10.1976 & -1.0992 & 0.0440  \\
        & GPT-4o-mini  &  0.1672 &	18.5799 &	0.1672 &	0.4879 &	57.8477 &	-0.0880 &	0.5752  \\
        & \textbf{Llama3.2-11B-Vis-Inst-finetuned (CTS1)}& \textbf{0.8958} & \textbf{89.5800} & \textbf{0.8958} & \textbf{0.9206} & \textbf{94.4300} & \textbf{0.6287} & \textbf{0.7573} \\
        & \textbf{Llama3.2-11B-Vis-Inst-finetuned (CTS1 + CTS2)}&0.8170 & 81.7045 & 0.8170 & 0.8202 & 91.5707 & 0.5667 & 0.7423 \\
        & \textbf{Llama3.2-11B-Vis-Inst-finetuned (CTS1 + CTS2 + CTS3)}&0.8184 & 81.8478 & 0.8184 & 0.8224 & 91.6472 & 0.5665 & 0.7433 \\ \midrule
        
        \multirow{7}{*}{\textbf{Class}}
        & Llama3.2-11B-Vision-Instruct &   0.3356 &	33.6372 &	0.3356 &	0.3351 &	60.1849 &	-0.2363 &	0.4422 \\
        & Qwen2.5-VL-7B-Instruct   &  0.8215 &	82.1546 &	0.8215 &	0.7081 &	88.6266 &	0.1243 &	0.6692  \\
        & MiniCPM-V-2-6     & 0.0000 & 0.0827 & 0.0000 & 0.0210 & 4.6216 & -1.2969 & 0.0109  \\
        & GPT-4o-mini &   0.7974 &	79.7497 &	0.7974 &	0.6844 &	89.2833 &	0.3698 &	0.6830  \\
        & \textbf{Llama3.2-11B-Vis-Inst-finetuned (CTS1)}&\textbf{0.9943} & \textbf{99.4300} & \textbf{0.9943} & 0.9696 & \textbf{99.8000} & 0.6056 & 0.8574 \\
        & \textbf{Llama3.2-11B-Vis-Inst-finetuned (CTS1 + CTS2)}&0.9937 & 99.3746 &0.9937 & \textbf{0.9712} & 99.7896 & \textbf{0.6101} & \textbf{0.8653} \\
        & \textbf{Llama3.2-11B-Vis-Inst-finetuned (CTS1 + CTS2 + CTS3)}&0.9935 & 99.3563 & 0.9935 & 0.9685 & 99.7913 & 0.6092 & 0.8518 \\ \midrule
        
        \multirow{7}{*}{\textbf{Packet}}
        & Llama3.2-11B-Vision-Instruct  &   0.0220 &	2.6485 &	0.0220 &	0.1539 &	26.2597 &	-0.6264 &	0.1512  \\
        & Qwen2.5-VL-7B-Instruct  &   0.0466 &	4.8793 &	0.0466 &	0.2385 &	35.6912 &	-0.6194 &	0.3911  \\
        & MiniCPM-V-2-6     & 0.0000 & 0.0651 & 0.0000 & 0.0150 & 6.6918 & -1.1320 & 0.0144 \\
        & GPT-4o-mini &   0.0525 &	5.7991 &	0.0525 &	0.2361 &	38.8444 &	-0.3666 &	0.3684 \\
        & \textbf{Llama3.2-11B-Vis-Inst-finetuned (CTS1)}&0.5890 & 58.9013 & 0.5890 & 0.7892 & 80.2759 & 0.6375 & 0.9994 \\
        & \textbf{Llama3.2-11B-Vis-Inst-finetuned (CTS1 + CTS2)}&0.9405 & 94.0543 & 0.9405 & \textbf{0.9616} & 97.1664 & \textbf{0.7972} & \textbf{0.9948} \\
        & \textbf{Llama3.2-11B-Vis-Inst-finetuned (CTS1 + CTS2 + CTS3)}& \textbf{0.9432} & \textbf{94.3204} & \textbf{0.9432} & 0.9535 & \textbf{98.5849} & 0.7397 & 0.9682 \\
        \bottomrule
    \end{tabular}
    \end{adjustbox}
    \caption{Preliminary results for Image2Code Task in the Continual Training Setup explored in the initial stage of the project}
    \label{tab:old_ablation}
\end{table*}

\end{document}